\documentclass[twoside]{article}

\usepackage[accepted]{aistats2025}

\usepackage[textsize=tiny]{todonotes}
\usepackage{mathtools}
\usepackage{multirow}
\usepackage{booktabs}
\usepackage{tablefootnote}
\usepackage{adjustbox}

\usepackage{caption}
\usepackage{subcaption}

\usepackage{tikz}
\usepackage{pgfplots}
\pgfplotsset{compat=1.17}
\usepackage{standalone}
\usetikzlibrary{shapes,arrows,fit,positioning, calc, decorations.markings}
\definecolor{captiongray}{RGB}{80,80,80}
\newcommand{\captioncomment}[1]{{\color{captiongray} \footnotesize #1}}

\usepackage{minitoc}

\usepackage[utf8]{inputenc} %
\usepackage[T1]{fontenc}    %
\usepackage{url}            %
\usepackage{booktabs}       %
\usepackage{amsfonts}       %
\usepackage{nicefrac}       %
\usepackage{microtype}      %

\usepackage{natbib}
\usepackage{multibib} %
\newcites{sup}{Supplementary References}
\setcitestyle{authoryear,round,citesep={;},aysep={,},yysep={;}}

\usepackage{amsfonts}
\usepackage{nicefrac}       %
\usepackage{amsmath}
\usepackage{amsthm}
\usepackage{amssymb}
\usepackage{cancel}
\usepackage{mathtools}
\usepackage{bbm}
\usepackage{bm}

\usepackage{cases}         %

\usepackage{mathalpha}

\usepackage{microtype}      %
\usepackage{enumitem}

\usepackage{pifont}

\usepackage{xspace}

\newcommand{\reals}{\mathbb{R}}

\newcommand{\defas}{\triangleq}

\newcommand{\T}{^{\top}} %

\newcommand{\vzero}{\boldsymbol{0}}

\newcommand{\vg}{\boldsymbol{g}}

\newcommand{\vu}{\boldsymbol{u}}

\newcommand{\vx}{\boldsymbol{x}}
\newcommand{\vy}{\boldsymbol{y}}

\newcommand{\vone}{\boldsymbol{1}}

\newcommand{\vtheta}{\boldsymbol{\theta}}
\newcommand{\vepsilon}{\boldsymbol{\epsilon}}

\newcommand{\Lag}{\mathfrak{L}}
\newcommand{\LagFL}{\Lag_{\mathrm{FL}}}
\newcommand{\LagRFL}{\Lag_{\mathrm{RFL}}}
\newcommand{\LERM}{L_{\text{\ERM}}}

\newcommand{\vlambda}{\boldsymbol{\lambda}}

\newcommand{\lrp}{\eta_{\vtheta}}
\newcommand{\lrd}{\eta_{\vlambda}}

\newcommand{\X}{\mathcal{X}}
\newcommand{\Y}{\mathcal{Y}}

\newcommand{\blobletter}[1]{\raisebox{.5pt}{\textcircled{\raisebox{-.8pt}{{\hspace{-1.1mm} \small #1}}}}}

\newcommand{\algo}[1]{{\small {\sffamily\textbf{#1}}}}
\newcommand{\ERM}{\algo{ERM}\xspace}
\newcommand{\CSERM}{\algo{CSERM}\xspace}
\newcommand{\FL}{\algo{FL}\xspace}
\newcommand{\Rawlsian}{\algo{Rawlsian}\xspace}
\newcommand{\RFL}{\algo{RFL}\xspace}

\newcommand{\FLeps}{\FL \hspace{-1.5ex} $(\vepsilon)$\xspace}

\newcommand{\SVMs}{\algo{SVMs}\xspace}

\definecolor{lightgray}{RGB}{230, 230, 230}
\definecolor{mathred}{RGB}{204, 69, 90}
\definecolor{mathblue}{RGB}{4, 78, 112}
\definecolor{mathgreen}{RGB}{1, 135, 70}

\definecolor{linkcolor}{RGB}{25,129,169}

\usepackage[colorlinks=true,
    linkcolor=linkcolor,
    citecolor=linkcolor,
    filecolor=linkcolor,
    urlcolor=linkcolor,
    pagebackref=true]{hyperref} 
\usepackage{etoolbox}
\makeatletter
\patchcmd\Hy@backout{\@auxout}{\@mainaux}{}{\fail}
\patchcmd\Hy@backout{\@auxout}{\@mainaux}{}{\fail} %
\makeatother

\usepackage{url}

\renewcommand*\backref[1]{\ifx#1\relax \else (Cit. on p. #1) \fi}

\usepackage[capitalize]{cleveref}

\Crefname{algorithm}{Algo.}{Algos.}
\Crefname{theorem}{Theorem}{Theorems}%
\Crefname{proposition}{Proposition}{Propositions}
\Crefname{appendix}{Appendix}{Appendices}%

\theoremstyle{plain}

\newtheorem{proposition}{Proposition}

\theoremstyle{remark}

\usepackage{enumitem}

\newcommand\blfootnote[1]{%
  \begingroup
  \renewcommand\thefootnote{}\footnote{#1}%
  \addtocounter{footnote}{-1}%
  \endgroup
}

\begin{document}

\runningauthor{Ramirez, Hounie, Elenter, Gallego-Posada, Hashemizadeh, Ribeiro, Lacoste-Julien}

\twocolumn[

\aistatstitle{Feasible Learning}

\aistatsauthor{ \hspace{2ex} Juan Ramirez$^{\star,1}$ \quad Ignacio Hounie$^{\star,2}$ \quad Juan Elenter$^{\star,3,\diamond}$  \quad  Jose Gallego-Posada$^{\star,1}$ \\
\vspace{2ex}  \hspace{4ex} \textbf{Meraj Hashemizadeh}$^{1}$ \qquad \textbf{Alejandro Ribeiro}$^{\dagger, 2}$ \qquad \textbf{Simon Lacoste-Julien}$^{\dagger, 1,4}$ }

\aistatsaddress{ \hspace{3ex} $^{1}$Mila \& Université de Montréal \quad  $^{2}$University of Pennsylvania \quad $^{3}$Spotify \quad $^{4}$Canada CIFAR AI Chair}

]

\begin{abstract}

We introduce Feasible Learning (\FL), a sample-centric learning paradigm where models are trained by solving a feasibility problem that bounds the loss for each training sample. In contrast to the ubiquitous Empirical Risk Minimization (\ERM) framework, which optimizes for average performance, \FL demands satisfactory performance \textit{on every individual data point}.
Since any model that meets the prescribed performance threshold is a valid \FL solution, the choice of optimization algorithm and its dynamics play a crucial role in shaping the properties of the resulting solutions. 
In particular, we study a primal-dual approach which dynamically re-weights the importance of each sample during training. To address the challenge of setting a meaningful threshold in practice, we introduce a relaxation of \FL that incorporates slack variables of minimal norm. Our empirical analysis, spanning image classification, age regression, and preference optimization in large language models, demonstrates that models trained via \FL can learn from data while displaying improved tail behavior compared to \ERM, with only a marginal impact on average performance.

\end{abstract}

\doparttoc %
\faketableofcontents %
\part{} %

\vspace{-9ex}
\section{INTRODUCTION}
\label{sec:intro}

Deep learning trends are shifting toward larger model architectures, as evidenced by GPT-4 \citep{openai2023gpt4}, DALLE-3 \citep{betker2023improving}, and Llama-3 \citep{dubey2024llama}. Larger models are capable of perfectly fitting increasingly large datasets, \textit{memorizing} the data by achieving near-zero loss on all samples \citep{arpit2017closer,zhang2017understanding}.
In this context, the Empirical Risk Minimization (\ERM) framework does not specify a preference among the many interpolating solutions. Consequently, the solution recovered in practice depends not only on the learning framework but also on the inductive biases of the chosen optimization algorithm. 
For instance, \citet{soudry2018implicit} highlight the role of stochastic gradient descent dynamics in guiding \ERM toward well-generalizing models.

While research has extensively focused on developing optimization algorithms suited for learning via \ERM \citep{kingma2015adam,gupta2018shampoo}, exploring alternative learning frameworks has received comparatively little attention.
Alternatives to \ERM could be better suited for specific learning scenarios, particularly when it is important to optimize for something other than average performance.
Such alternatives may exhibit distinct properties, such as improved uncertainty quantification \citep{balasubramanian2014conformal}, fairness \citep{lahoti2020fairness}, or robustness \citep{mkadry2017towards}. 

Given the abundance of interpolating, well-generalizing solutions in modern machine learning problems, why would we limit ourselves to those derived from \ERM?

\vspace{-4ex}

\blfootnote{\hspace{-5ex} $^\star$Equal contribution. $^\dagger$Equal supervision. \\
$^\diamond$Work done while at the University of Pennsilvania. \\
Correspondance to: \href{juan.ramirez@mila.quebec}{\texttt{juan.ramirez@mila.quebec}}}

\begin{figure*}[ht]
    \centering
    \includegraphics{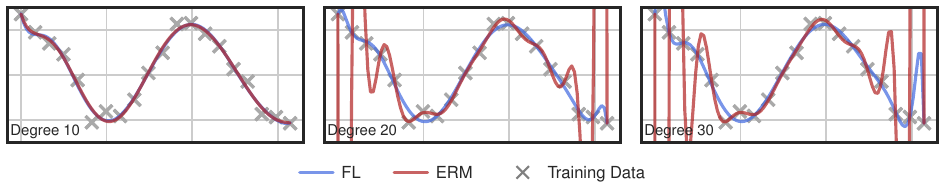}
    \vspace{-1ex}
    \caption{Fitting polynomials of varying degrees: \textbf{While \ERM tends to overfit with high-degree polynomials, \FL still recovers smoother solutions}. 
    This occurs even though non-smooth solutions are also part of the \FL solution set, highlighting the influence of the optimization algorithm on the final solution.
    \captioncomment{
    Due to ill-conditioning, we solve both problems using a standard convex optimization solver. The data is generated by adding Gaussian noise ($\sigma=0.2$) to a cosine wave. \FL's constraint value is set to one standard deviation, $\epsilon = \sigma$. See \Cref{app:details} for details. 
    }
    }
    \label{fig:1}
    \vspace{-2ex}
\end{figure*}

In this paper, we introduce a novel learning framework called Feasible Learning (\FL, \S\ref{sec:feasible_learning}). \FL formulates learning as a \textit{feasibility} problem, where we seek a predictor that meets a \emph{bounded loss constraint for all training samples}. Unlike the ubiquitous \ERM framework, which optimizes for average performance, \FL demands a minimum performance level for each data point.

Concretely, \FL formulates an optimization problem with a trivial (constant) objective function, while imposing a constraint on the loss of the predictor~\smash{$h : \mathcal{X} \to \mathcal{Y}$} for each training sample ${(\vx_i, \vy_i)}_{i=1}^n$:
\begin{equation}
    \label{eq:FL_intro}\tag{\FL}
    \hspace{2mm} \min_{h \in \mathcal{H}} \, 0 \hspace{3mm} \text{s.t} \hspace{3mm} \ell(\vy_i, h(\vx_i)) \le \epsilon \hspace{2mm} \text{for} \hspace{2mm} i=1, \ldots, n,
\end{equation}
where $\epsilon \geq 0$ is the maximum allowed per-sample loss.

The main properties of \FL problems are presented in \S\ref{sec:feasible_learning} and can be summarized as follows: \blobletter{1} \FL is inherently \textit{sample-centric}, requiring satisfactory performance across all training samples. This contrasts with the \ERM framework, which aims to optimize average performance and may overlook poor performance on individual samples. \blobletter{2} \FL does not establish a preference between feasible models that meet the constraints: while \ERM seeks solutions with zero training loss, \FL accepts any solution that satisfies the minimum performance threshold $\epsilon$ for each data point. 
This leads to \blobletter{3} \FL inducing functional regularization of the loss function which prevents overfitting to the training data.\footnote{Provided the optimization algorithm of choice does not incentivize reducing the per-sample loss beyond $\epsilon$.}

\Cref{fig:1} illustrates the behavior of \FL on a polynomial regression task. Both \ERM and \FL find solutions that fit the data well. 
Note that the \ERM solution is also a valid solution to the \FL problem. However, \FL recovers a qualitatively different predictor.
Whereas the existence of such non-interpolating solutions is a property of the \FL problem, finding them in practice depends on the choice of optimization algorithm and its dynamics.

We adopt a primal-dual optimization approach to solve \FL problems, which amounts to performing a weighted version of \ERM (\S\ref{sec:solving_fl}). The weight of each sample corresponds to the Lagrange multiplier associated with its constraint, which is dynamically adjusted based on the ``difficulty'' of fitting said sample.
We favor this approach because \blobletter{1} the algorithmic similarity with \ERM allows to leverage established techniques for training deep neural networks, \blobletter{2} enabling it to scale to high-dimensional problems with large numbers of constraints; and \blobletter{3} it does not inherently favor interpolating solutions over non-interpolating ones.

A challenging aspect of \FL is determining a suitable constraint level $\epsilon$. If set too tightly, the \FL problem may be infeasible, i.e. it may not admit any solutions. Conversely, setting $\epsilon$ too loosely may fail to ensure good performance. 
To address this, we propose relaxing the \FL problem with slack variables of minimal norm that would make the problem feasible.
We call this approach Resilient Feasible Learning (\RFL, \S\ref{sec:resilient_fl}).
Our primal-dual approach to solve \RFL problems (\S\ref{sec:solving_rfl}) enjoys enhanced convergence guarantees compared to primal-dual \FL. 

To gain a deeper understanding of Feasible Learning, we tackle the following questions: \\
\hspace*{2mm} (\textbf{Q1}) \underline{Can we learn via \FL}? Yes, deep networks trained via \FL achieve comparable average performance to \ERM on the train and test sets (\S\ref{exps:learning}), with equivalent training cost and similar hyperparameter robustness. \\
\hspace*{2mm} (\textbf{Q2}) \underline{How does \RFL help}? In problems where infeasibility leads to poor optimization dynamics for \FL, \RFL alleviates this issue, achieving good performance (\S\ref{exps:dynamics}). \\
\hspace*{2mm} (\textbf{Q3}) \underline{How do \FL solutions compare to \ERM}? We observe that \FL produces a more concentrated loss distribution across (training and test) samples, resulting in fewer instances with excessively high losses (\S\ref{exps:robustness}). %

The main contributions of our work are as follows:\footnote{Our code is available at: \texttt{\url{https://github.com/juan43ramirez/feasible-learning}}.}
\begin{itemize}
    \vspace{-2ex}
    
    \item We propose Feasible Learning (\FL), framing learning as a constraint satisfaction problem (\S\ref{sec:feasible_learning}).

    \vspace{-1ex}
    \item We introduce Resilient Feasible Learning (\RFL, \S\ref{sec:resilient_fl}), a relaxation of \FL that addresses potential infeasibility issues. 
    We show that \RFL is equivalent to a non-convex, strongly-concave min-max problem (\Cref{prop:strong_duality}). %

    \vspace{-1ex}
    \item We provide primal-dual algorithms for solving \FL (\S\ref{sec:solving_fl}) and \RFL (\S\ref{sec:solving_rfl}) problems, which are as cheap as gradient descent on the \ERM objective (up to the negligible cost of updating the dual variables). 

    \vspace{-1ex}
    \item We perform an empirical exploration of \FL and \RFL problems and their solutions (\S\ref{sec:experiments}), supporting our answers to questions \textbf{Q1}-\textbf{Q3}.

    \vspace{-2ex}
\end{itemize}

\textbf{Scope:} We introduce \FL, present a practical algorithm for solving it, and provide empirical evidence that \FL is a compelling alternative to the widely used \ERM framework. However, we do not intend to claim that \FL outperforms \ERM or other learning paradigms. As with any multi-objective problem, adequately balancing competing objectives (such as the losses on each sample) depends on the specific application and requirements.

Developing a statistical decision theory framework for analyzing \FL problems is beyond the scope of our work. In particular, future research should determine which statistical goals \FL problems are best suited for. %
Rather, we explore \FL empirically and investigate its properties.

\section{FEASIBLE LEARNING}
\label{sec:feasible_learning}

We consider the problem of learning a predictor \smash{$h_{\vtheta}: \X \rightarrow \Y$} with parameters $\vtheta \in \Theta$ on a labeled dataset $\mathfrak{D} = \{(\vx_i, \vy_i)\}_{i=1}^n$.  The quality of the predictions is measured by a differentiable (surrogate) loss function \smash{$\ell : \Y^2 \rightarrow \reals_{\geq 0}$}. To simplify the notation, we denote the loss incurred on the $i$-th data point as $g_i(\vtheta) \defas \ell(\vy_i, h_{\vtheta}(\vx_i))$, and the vectorized version of these losses as \smash{$\vg(\vtheta) = [g_1(\vtheta), \ldots, g_n(\vtheta)]^\top$}.

The Feasible Learning (\FL) paradigm advocates for learning through solving a \textit{feasibility problem}. Specifically, \FL considers an optimization problem with a trivial, constant objective while enforcing a loss constraint for each training sample:
\begin{equation}
    \vspace{-1ex}
    \label{eq:FLeps}\tag{\FLeps}
    \hspace{2mm} \min_{\vtheta \in \Theta} \,\, 0 \hspace{3mm} \text{s.t} \hspace{3mm} \vg(\vtheta) \le \vepsilon, \hspace{2mm} 
\end{equation}
where $\vepsilon = \epsilon \vone$ is the constraint level.\footnote{Using different constraint levels per sample is possible.} 
In the \FL framework, a model is acceptable \textit{if and only if} it achieves a sufficiently small loss (given $\epsilon$) on every training point. 
Choosing $\ell$ as the squared error imposes a bound on the maximum error per sample, while upper-bounding the per-sample cross-entropy sets a lower bound on the probability the model assigns to the correct label.

When the model class $\Theta$ can interpolate the training data, setting $\epsilon = 0$ results in the set of solutions for \FL matching that of \ERM\hspace{-1ex}—consisting of those that achieve zero training loss on all samples. When $\epsilon > 0$, \FL admits additional solutions—those that satisfy the constraints but do not necessarily interpolate the data. 

In the non-interpolation case, we expect \FL to outperform \ERM in terms of the maximum loss over the dataset---potentially at the expense of a heightened average loss. However, this trend may not always manifest in practice due to the inherent challenges of solving non-convex (constrained) optimization problems.

\textbf{Functional regularization.}
\FL does not inherently favor interpolating or non-interpolating solutions, as long as both satisfy the constraints. This introduces a form of \textit{functional} regularization: to prevent excessive minimization of the loss, \FL regularizes \textit{the loss itself} by not demanding to reduce it beyond a specified threshold $\epsilon$. This approach to regularization is model-agnostic, differing from standard techniques that rely on objectives like smoothness, small norms, or sparsity.

When the model class $\Theta$ cannot interpolate the data, certain values of $\epsilon$ may result in infeasible \FL problems—this can occur in tasks with low model capacity or with noisy or mislabeled data. This introduces a trade-off when selecting an appropriate $\epsilon$: tight values can result in infeasible problems, while overly loose values may lead to vacuous constraints, potentially failing to ensure good performance. As appropriate constraint levels depend on the model parametrization, data, and task, selecting them can be challenging in practice.

While our experiments demonstrate that \FL can recover useful solutions even if they do not satisfy all constraints (\S\ref{exps:learning}), it remains desirable to make the \FL framework robust against problem misspecification. Thus, in \S\ref{sec:resilient_fl} we propose finding the minimum norm constraint relaxation needed to make the problem feasible.

\vspace{-1ex}
\subsection{Solving \FL Problems}
\label{sec:solving_fl}

Even if the loss function $\ell$ is convex in its inputs \smash{$(\vy_i, h_{\vtheta}(\vx_i))$}, it may not be convex in~$\vtheta$. Therefore, \smash{\ref{eq:FLeps}} is typically a \textit{non-convex} constrained optimization problem with no closed-form solution. 
Furthermore, the lack of an objective function and the non-convexity of the feasible set preclude the use of standard constrained optimization techniques such as projected gradient descent \citep{goldstein1964convex} or Frank-Wolfe \citep{frank1956algorithm}. Instead, we leverage Lagrangian duality. The min-max Lagrangian game associated with \ref{eq:FLeps} is:
\begin{equation}
    \label{eq:FL_lagrangian}
    \min_{\vtheta \in \Theta}\, \max_{\vlambda \ge \vzero} \, \LagFL(\vtheta, \vlambda) \defas  \vlambda\T (\vg(\vtheta) - \vepsilon),
\end{equation}
where $\vlambda \geq \vzero$ is the vector of Lagrange multipliers associated with the constraints.
We refer to $\vtheta$ as the \textit{primal} variables, and $\vlambda$ as the \textit{dual} variables.  
\FL yields a Lagrangian with one multiplier $\lambda_i$ per datapoint $\vx_i$.

A simple algorithm for finding min-max points of $\LagFL$ is to perform gradient descent steps on $\vtheta$ and projected gradient ascent steps on $\vlambda$ \citep[GDA]{arrowhurwitz}. Alternating GDA updates \citep{zhang2022near} yield:
\begin{align}
    \vspace{-2ex}
    \begin{split}
    \label{eq:sim_gda_FL}
        \vlambda_{t+1} 
        &\leftarrow
        \Big[ \vlambda_t + \lrd \underbrace{\left( \vg(\vtheta_t) - \vepsilon \right)}_{\nabla_{\vlambda} \LagFL(\vtheta_t, \vlambda_t)} \Big]_+ \\
        \vtheta_{t+1} 
        & \leftarrow \vtheta_t - \lrp \bigg[ \underbrace{\sum_{i=1}^n \vlambda^{(i)}_{t+1} \, \nabla_{\vtheta} \, g_i(\vtheta_t)}_{\nabla_{\vtheta} \LagFL(\vtheta_t, \vlambda_{t+1})} \bigg],
    \end{split}
\end{align}
where $[\, \cdot \,]_+$ denotes a projection onto $\reals_{\geq \vzero}^n$ to enforce $\vlambda \geq 0$, and $\eta_{\{\vx,\vlambda\}}$ are step sizes. We initialize $\vlambda_0 = \vzero$.

The primal updates in \cref{eq:sim_gda_FL} resemble gradient descent on the \ERM objective by following the gradients of the per-sample losses. However, unlike \ERM, where these gradients are weighted equally, \FL uses the Lagrange multipliers as weights. Since these multipliers are \textit{optimized}, the algorithm dynamically re-weights the importance of each data point throughout training.

The re-weighting works as follows: if $g_i(\vtheta) > \epsilon$, the corresponding multipler increases; if $g_i(\vtheta) < \epsilon$, the multiplier decreases, potentially reaching zero. 
Consequently, data points with consistently high losses result in large multipliers, causing the primal updates to focus on reducing their loss.
Conversely, data points with consistently small losses have small or even zero multipliers, allowing them to be largely ignored during optimization.
Thus, a given primal update is influenced by the ``instantaneous'' incentive to satisfy a constraint, reflected in the loss gradient, and the ``historical difficulty'' of satisfying the constraint, captured in the multiplier. In \S\ref{exps:robustness}, we show how hard samples—such as mislabeled ones—tend to yield large multipliers.

\textbf{Functional regularization.}
These optimization dynamics do not aim to satisfy the constraints beyond the prescribed level $\epsilon$. Once a constraint is strictly satisfied, the dual updates reduce the corresponding multiplier, discouraging the primal updates from further minimizing the loss for the corresponding sample.

However, satisfying the constraint for some samples may require achieving a loss tighter than $\epsilon$ on others. Additionally, primal-dual methods can overshoot into the interior of the feasible set due to a ``delay'' between initially meeting the constraint and sufficiently reducing the multiplier to relieve pressure on loss reduction. This overshoot can be mitigated by using PI controllers to update the multipliers instead of relying on gradient ascent \citep{stooke2020responsive,sohrabi2024pi}.

\textbf{Infeasible problems.}
When applying GDA to infeasible problems, the multipliers associated with unsatisfiable constraints will increase indefinitely. This can potentially lead to numerical instability, disrupting optimization. However, this issue does not arise in our proposed method for solving \RFL problems (\S\ref{sec:solving_rfl}).

\textbf{The cost of GDA on \FL.}
Since the primal update direction \(\nabla_{\vtheta} \LagFL\) is a linear combination of the per-datapoint loss gradients \(\nabla_{\vtheta} g_i(\vtheta)\), it can be computed efficiently using automatic differentiation, without needing to store each gradient individually. This makes its computation as efficient as that of the \ERM loss gradient. Therefore, applying (mini-batch) gradient descent-ascent on $\LagFL$ is as efficient as performing (mini-batch) gradient descent on the \ERM loss up to the cost of storing and updating the multipliers.
This overhead is negligible when the $\text{dim}(\vtheta)$ is much larger than $n$.

\textbf{Practical remarks.}
More advanced techniques than GDA are often used for solving Lagrangian min-max problems. Standard deep learning optimizers like Adam \citep{kingma2015adam} can be used for the primal updates, while recent work suggests using PI control to enhance the optimization dynamics of the multipliers \citep{stooke2020responsive,sohrabi2024pi}.

We favor alternating GDA over simultaneous GDA as it offers better convergence guarantees under primal strong convexity \citep{zhang2022near} without incurring additional computational cost \citep{sohrabi2024pi}. 
This approach is particularly relevant for \FL, where the initial primal update has no effect due to the initialization $\vlambda_0 = \vzero$.
Thus, it is essential to first ``warm up'' $\vlambda$, which is naturally achieved by using an alternating scheme that updates the dual variables first.

Machine learning tasks typically involve computations over mini-batches of data. \FL supports these by: \blobletter{1} sampling a mini-batch and computing the losses for each sample, \blobletter{2} updating the multipliers for the observed samples, and \blobletter{3} performing a stochastic gradient step for the model. This results in stochastic updates on $\vtheta$ and coordinate-wise updates on $\vlambda$.

\section{\hspace{-1.4ex} RESILIENT FEASIBLE LEARNING}
\label{sec:resilient_fl}

The potential misspecification of \FLeps problems can be addressed by relaxing the constraints using \textit{slack variables}, denoted by $\vu$.\footnote{Similar to soft-margin support vector machines.} Given $\alpha > 0$, we consider the following constrained optimization problem:
\vspace{-1ex}
\begin{equation}
    \tag{\RFL\hspace{-1.2mm}$(\vepsilon, \alpha)$} 
    \label{eq:rfl}
    \hspace{2mm} \min_{\vtheta \in \Theta, \vu \ge \vzero} \, \frac{\alpha}{2} ||\vu||^2 \hspace{3mm} \text{s.t} \hspace{3mm} \vg(\vtheta) \le \vepsilon + \vu. 
    \vspace{-1ex}
\end{equation}
We call this approach Resilient Feasible Learning (\RFL) due to its robustness to problem misspecification. If the original \FL problem is feasible, the corresponding \RFL problem is equivalent, as the optimal relaxation $\vu$ will be zero. Crucially, \RFL \textit{guarantees the existence of a feasible solution}, even when the original \FL problem is infeasible.
We generally favor \RFL over \FL in practice since it alleviates the challenge of setting $\epsilon$.

In \ref{eq:rfl}, $u_i > 0$ represents a strict relaxation of the $i$-th constraint, while $u_i = 0$ indicates that it remains unchanged. The cost of relaxing constraints depends on the norm of the slack variables.
Although other norms could be used, we focus on the $\text{L}_2$-norm due to its algorithmic advantages, as demonstrated in Prop.~\ref{prop:strong_duality}.

While one might consider setting $\epsilon = 0$--thus allowing \RFL to determine the tightest possible loss requirements through the slacks--maintaining a positive $\epsilon$ enables regularization, as in \FL. 
This prevents the model from overly minimizing per-sample losses, even if \textit{some} data points are allowed to violate the constraints. For example, $\epsilon$ could demand marginally correct predictions, while $\vu$ allows mislabeled samples to be misclassified.

Although the parameter $\alpha$ does not change the optimal solution $\vtheta^*, \vu^*$ of \ref{eq:rfl}, it can affect the dynamics of the algorithm used to solve it. 
We present a formulation with an arbitrary $\alpha$ to allow flexibility in tuning these dynamics. Its effect is explored in \cref{table:ablation_main}.

\vspace{-1ex}
\subsection{Solving \RFL Problems}
\label{sec:solving_rfl}

As with \FL problems, we use the Lagrangian approach to solve \RFL problems. The min-max Lagrangian game associated with \ref{eq:rfl} is given by:
\vspace{-1ex}
\begin{equation}
    \label{eq:FLS_lagrangian}
    \min_{\vtheta \in \Theta, \vu \ge \vzero} \, \max_{\vlambda \ge \vzero} \,\, \underbrace{\frac{\alpha}{2} ||\vu||^2 + \vlambda\T (\vg(\vtheta) - \vepsilon - \vu)}_{\LagRFL(\vtheta, \vu, \vlambda)}
    \vspace{-1ex}
\end{equation}
We now transform \cref{eq:FLS_lagrangian} to a problem without slacks.
\begin{proposition}
    \label{prop:strong_duality}
    [\hyperref[app:proofs:prop1]{Proof}] For every $\vtheta \in \Theta$, the following strong duality condition holds:
    \vspace{-1ex}
    \begin{align}
        \hspace{-1.6ex} \min_{\vu \ge \vzero}  \max_{\vlambda \ge \vzero} \LagRFL(\vtheta, \vu, \vlambda) 
        &= \max_{\vlambda \ge \vzero}  \min_{\vu \ge \vzero} \LagRFL(\vtheta, \vu, \vlambda) \\
        &=  \max_{\vlambda \ge \vzero}  \underbrace{\vlambda\T \hspace{-0.5ex} (\vg(\vtheta) \hspace{-0.3ex} - \hspace{-0.3ex}\vepsilon)}_{\LagFL(\vtheta,\vlambda)} - \frac{||\vlambda||^2}{2 \alpha} 
    \end{align}
    \vspace{-2ex}
\end{proposition}
As a consequence of \Cref{prop:strong_duality}, the Lagrangian problem for \RFL in \cref{eq:FLS_lagrangian} can be solved via a quadratically-regularized version of the \FL Lagrangian:
\begin{equation}
\vspace{-1ex}
\label{eq:quadratically_regularized_lagrangian}
    \min_{\vtheta \in \Theta} \, \max_{\vlambda \ge \vzero} \, \, \Lag_{\alpha}(\vtheta, \vlambda) 
    \defas
    \LagFL(\vtheta, \vlambda)  - \frac{1}{2 \alpha} ||\vlambda||^2.
\end{equation}
$\Lag_{\alpha}$ is strongly concave on $\vlambda$, implying that for a fixed $\vtheta$, the inner maximization has a unique solution $\vlambda^*$ (whereas \FL may yield an unbounded inner problem). 

This formulation is advantageous, as gradient descent-ascent offers convergence guarantees for non-convex, strongly-concave min-max problems \citep{lin2020gradient}. In particular, strong convexity of the function $\vg$ yields a linear convergence rate \citep{chen1997convergence}.

Alternating GDA updates on $\Lag_{\alpha}$ yield similar updates to those of primal-dual \FL (\cref{eq:sim_gda_FL}).
The primal update direction remains the same: a linear combination of the per-sample loss gradients, weighted by the multipliers. The dual update includes a weight decay of $1/\alpha$, which “discounts” historical violations, resulting in different dynamics. For example, this prevents the multipliers for unsatisfiable constraints from growing indefinitely.

By analytically solving the inner maximization problem in \cref{eq:quadratically_regularized_lagrangian}, we recover the following result:

\begin{proposition}
    \label{prop:clamped_quadratic}

    [\hyperref[app:proofs:prop2]{Proof}]
    For every $\vtheta \in \Theta$, we have:
    \begin{equation}
        \label{eq:clamped_quadratic_erm}
        \min_{\substack{\vtheta \in \Theta \\ \vu \ge \vzero}} \, \max_{\vlambda \ge \vzero}\,\,  \LagRFL(\vtheta, \vu, \vlambda) 
        = 
        \min_{\vtheta \in \Theta} \,\, \frac{\alpha}{2} \left\Vert \left[ \vg(\vtheta) \hspace{-0.3ex} - \hspace{-0.3ex} \vepsilon \right]_+ \right\Vert^2
    \end{equation}
\end{proposition}

Therefore, \ref{eq:rfl} can be solved via either \blobletter{1} a non-convex, strongly-concave min-max problem (\cref{eq:quadratically_regularized_lagrangian}), or \blobletter{2} a non-convex \ERM-style minimization problem with a clamped-and-squared loss (\CSERM, \cref{eq:clamped_quadratic_erm}). 
While these two problem formulations are equivalent, we favor the primal-dual approach due to its optimization dynamics, which are explored in \S\ref{exps:dynamics}.

\section{RELATED WORK}

\textbf{Learning paradigms}. \FL stands in contrast to the standard Empirical Risk Minimization (\ERM):
\begin{equation}
    \label{eq:ERM}\tag{\ERM}
    \hspace{2mm} \min_{\vtheta \in \Theta} \,  \LERM(\vtheta) \defas \frac{1}{n} \vone\T \vg(\vtheta), \hspace{2mm}
\end{equation}
which views the learning problem as ``choosing from the given set of functions the one which approximates best the supervisor's response'' \citep[p.2]{vapnik1991erm}. 
\ERM operationalizes the notion of ``best approximation'' through the \textit{average} loss across the training set.

There is a fundamental difference between the goals of the \FL and \ERM problems. 
To illustrate this, consider the case of recommender systems used by streaming or social media platforms. Service providers often prioritize metrics like average click-through rates or watch-time to measure overall system success and user engagement. However, individual users are primarily concerned with how well the recommendations align with \textit{their} personal tastes and preferences. A system that performs well on average might still fail individual users by consistently suggesting irrelevant or inadequate content.
\ERM inherently allows for trade-offs between training samples, allowing models to perform poorly on certain samples as long as they compensate by performing exceptionally well on others. %

In contrast to robust (\Rawlsian) approaches \citep{lahoti2020fairness}, which minimize the worst-case risk:
\begin{equation}
    \label{eq:rawlsian}
    \min_{\vtheta \in \Theta} \,\, \max_{i \in \{1,\dots,n\}} \,\, g_i(\vtheta), \tag{\Rawlsian}
\end{equation}
\FL only requires that the upper bound in the per-sample loss is satisfied\footnote{The \Rawlsian approach ``finds'' the tightest $\epsilon$ that would ensure feasibility in a corresponding \FL problem.}. Thus, \FL does not prefer one model over another as long as both satisfy the constraints. 

The pursuit of minimizing the average loss in \ERM or the maximum loss in the \Rawlsian approach can lead to models that overfit the training data or become overly confident. This excessive reduction in losses can harm generalization, motivating the use of regularization techniques. Unlike traditional methods, which promote parsimony using surrogate criteria like L$_p$-norms, \FL explicitly establishes an upper bound \textit{on the loss itself} through the constraint level $\epsilon$. %

\textbf{Learning through constraints.}
The use of data samples to constrain the parameter space has long been applied in generative modeling and parametric estimation, dating back at least to the Maximum Entropy principle \citep{Jaynes1957Maxent}. However, moment-constrained approaches like rate-constrained classification \citep{Neyman-PearsonClassification} rely on aggregate statistics of the samples. In contrast, \FL considers constraints on the samples.

\textbf{\SVMs.}
Hard-margin Support Vector Machines (\SVMs) find classifiers that \blobletter{1} correctly classify all points and \blobletter{2} do so with the maximum possible margin. In contrast, \FL does not explicitly seek the maximum margin. However, it can be interpreted as finding a classifier that guarantees a certain confidence on the surrogate loss $\ell$, as determined by $\epsilon$. Moreover, our primal-dual approach allows \FL to find meaningful solutions even when the problem is infeasible, whereas hard-margin \SVMs are ineffective for non-separable data.

\textbf{Constrained \ERM.} The standard approach in constrained machine learning typically adds constraints to standard training objectives (such as the average loss) to enforce requirements such as fairness \citep{cotter2019proxy}, sparsity \citep{gallego2022controlled}, or safety \citep{stooke2020responsive}. Thus, constraints are often used to encourage behaviors that drift away from the main learning objective. In contrast, \FL considers constraints as the primary driving force for learning.

\textbf{Resilience.}
Previous work has addressed constraint-level misspecification in constrained \ERM. \citet{hounie2024resilient} propose Resilient Constraint Learning, which, like \RFL, uses slack variables to relax constraints. Their method relaxes constraints based on the \textit{sensitivity} of the objective function to each constraint (i.e., the potential improvement in the objective given a small relaxation). 
In the case of \FL, however, the objective is constant, so there is no trade-off between a learning objective and the imposed constraints to consider.

\textbf{Clamped losses.}
Using a thresholded loss for regression—where errors below a certain threshold are not penalized—dates back at least to \citet[Chapter 6]{vapnik1998statistical}. However, this approach still allows the model to trade off errors between data points that exceed the threshold. In contrast, \FL does not permit infeasible solutions, thereby preventing such trade-offs. 
On the other hand, \RFL is equivalent to \ERM using a thresholded (and also squared) loss (see \cref{prop:clamped_quadratic}).
Unlike standard thresholded loss approaches, \RFL does not regularize the model's complexity.

\section{EXPERIMENTS}
\label{sec:experiments}

In this section, we empirically evaluate the Feasible Learning framework, demonstrating that \FL and \RFL present a compelling alternative to the widely used \ERM framework.
We demonstrate that models trained via \FL can learn (\S\ref{exps:learning}).
We also explore the advantages of \RFL over \FL (\S\ref{exps:dynamics}) and analyze their loss distribution profiles (\S\ref{exps:robustness}).
See \Cref{app:details} for details on our experimental setup.
For comprehensive results, see \Cref{app:experiments}.

\textbf{Tasks.} 
We train ResNet-18 models \citep{he2016deep} for CIFAR10 \citep{cifar} classification and for UTKFace \citep{utkface} age regression. 
We also fine-tune an 8 billion parameter Llama-3.1 model~\citep{dubey2024llama} on a cleaned version of Intel Orca DPO pairs dataset.\footnote{\texttt{\url{https://huggingface.co/datasets/argilla/distilabel-intel-orca-dpo-pairs}}} We use Direct Preference Optimization (DPO)~\citep{rafailov2024dpo}. %
Finally, we train a Multi-Layer Perceptron for Two-Moons classification.
\Cref{tab:expt-details} in App.~\ref{app:details} lists each task's training set size, which corresponds to the number of constraints.

As the constraint level $\epsilon$ is expressed in terms of the loss, it can be interpreted for each task. 
\textbf{Classification}: we bound the cross-entropy loss, which translates into a lower bound on the predicted probability for the true class.\footnote{$\hat{p}_{\text{true}} \geq \exp(-\epsilon)$, where $\hat{p}_{\text{true}}$ is the model’s predicted probability for the correct label.} 
\textbf{Regression}: we bound the Squared Error (SE), which corresponds to the difference in years between the predicted and true ages. We normalize the ages to have zero mean and unit variance.
\textbf{Preference Alignment (DPO)}: The DPO loss constraint is expressed as $\sigma(r(y^{+}) - r(y^{-})) \geq \exp(-\epsilon)$, where $y^{+}$ and $y^{-}$ represent a pair of preferred and dispreferred completions, respectively, $r$ is an implicit reward model defined via log-likelihood ratios, and $\sigma$ is a sigmoid function.

\textbf{Methods.} We train models via \blobletter{1} \ERM, \blobletter{2} \CSERM: Clamped-and-Squared \ERM (\cref{eq:clamped_quadratic_erm}), \blobletter{3} \FL: Feasible Learning, and \blobletter{4} \RFL: Resilient Feasible Learning.

\textbf{Experimental uncertainty.} 
Unless stated otherwise, all reported metrics are averaged over 5 seeds.

\textbf{Software \& Hardware.} Our implementations use PyTorch \citep{pytorch} and the Cooper library for constrained optimization \citep{gallegoPosada2024cooper}. Experiments are run on NVIDIA \texttt{L40S} GPUs.

\vspace{-0.5ex}
\subsection{Can We Learn with Feasible Learning?}
\label{exps:learning}

We begin by evaluating models trained with \FL using \ERM's primary success criterion: average performance. 
Despite \FL tackling a different problem—and irrespective of its effectiveness in solving it—we assess whether \FL still succeeds in the standard learning task.

\begin{table}[t!]
\centering
\caption{Final performance for CIFAR10. \textbf{\FL and \RFL achieve comparable average losses and accuracies to \ERM, on both the training and test sets.}}
\label{table:main}
\resizebox{0.49\textwidth}{!}{
\begin{tabular}{lccccc}
\toprule
\multirow{2}{*}[-0.5\dimexpr \aboverulesep + \belowrulesep + \cmidrulewidth]{\textbf{Method}} & \multirow{2}{*}[-0.5\dimexpr \aboverulesep + \belowrulesep + \cmidrulewidth]{\textbf{$\epsilon$}} & \multicolumn{2}{c}{\textbf{Train}} & \multicolumn{2}{c}{\textbf{Test}} \\
\cmidrule(lr){3-4} \cmidrule(lr){5-6}
\multicolumn{1}{c}{} & \multicolumn{1}{c}{} & \multicolumn{1}{c}{CE Loss} & \multicolumn{1}{c}{Acc.} & \multicolumn{1}{c}{CE Loss} & \multicolumn{1}{c}{Acc.} \\
\midrule
\ERM &  & 0.00 & 1.00 & 0.30 & 0.93 \\
\midrule
\FL & $0.51$ & 0.01 & 1.00 & 0.34 & 0.92 \\
\RFL & $0.51$ & 0.01 & 1.00 & 0.35 & 0.92 \\
\CSERM & $0.51$ & 0.34 & 0.98 & 0.52 & 0.88 \\
\midrule
\FL & $0.00$ & 0.00 & 1.00 & 0.33 & 0.93 \\
\RFL & $0.00$ & 0.00 & 1.00 & 0.34 & 0.93 \\
\CSERM & $0.00$ & 0.07 & 0.99 & 0.42 & 0.87 \\
\bottomrule
\end{tabular}
}

\end{table}

\Cref{table:main} presents results for a CIFAR10 classification task. We include \FL and \RFL under two requirements: $\epsilon=0$, where the model is required to assign a probability of $1$ to the correct label, matching \ERM's solution set assuming interpolation is possible; and $\epsilon=0.51$, where a true class probability of $0.6$ is required, ensuring correct classification with a small margin.

These results demonstrate that \FL and \RFL can effectively learn classifiers with only a slight degradation in average performance compared to \ERM on both the training and test sets. However, \FL and \RFL may offer advantages in tail behavior and robustness (\S\ref{exps:robustness}), making this trade-off appealing for certain applications. We observe this trend across all tasks (see \Cref{app:experiments}).

\textbf{Optimization budget.}
Notably, \FL and \RFL achieve comparable performance to \ERM \textit{within the same training budget} of 200 epochs. However, it is important to note that poor choices of the dual step size can cause \FL and \RFL to converge more slowly if chosen too small, or experience degraded performance if set too high.

\textbf{Robustness.}
We found that, despite introducing a new hyper-parameter with the dual step size, \FL and \RFL are \blobletter{1} similarly robust to the choice of the primal step size as \ERM, and \blobletter{2} fairly robust to the choice of the dual step size, achieving good performance across multiple orders of magnitude (see \Cref{app:experiments}).

\vspace{-1ex}
\subsection{How Does Resilience Help?}
\label{exps:dynamics}

\Cref{table:ablation_main} presents an ablation study on the choice of \RFL's $\alpha$ for the UTKFace age regression task. We select $\epsilon=0.0$—which is unattainable due to the presence of duplicated samples in the dataset with different labels—to emphasize the benefits of resilience in providing flexibility to satisfy the constraints.

\FL's constraints are too restrictive, leading to poor average and maximum performance, significantly worse than \ERM. We attribute this to its optimization dynamics, which cause some multipliers to grow indefinitely, destabilizing the optimization process. In contrast, \RFL can relax these requirements and achieve performance comparable to \ERM. In particular, \RFL($\alpha=10^{-3}$) outperforms \ERM in average train and test errors.
Moreover, although \RFL relaxes the constraints, potentially allowing for larger maximum errors than \FL, it achieves smaller Max SE's, further indicating a failure of \FL.

Moreover, we observe that while certain values of $\alpha$ may yield better performance, a wide range of values spanning multiple orders of magnitude can still result in strong performance. In other words, RFL demonstrates relatively low sensitivity to $\alpha$.

A trade-off in using \RFL is that, even though the choice of $\epsilon$ becomes less critical, we now need to select an appropriate $\alpha$. 
Our findings across various choices of $\epsilon$ and tasks indicate that finding suitable $\alpha$ values may require extensive tuning (see \cref{app:experiments}). 

\begin{table}[t!]
\centering
\caption{Final performance for UTKFace age regression. \FL's constraints are too restrictive, resulting in worse performance than \ERM. \textbf{In contrast, \RFL can outperform \ERM with appropriate $\alpha$ choices}. \captioncomment{$\epsilon=0.0$. Max SE stands for Maximum per-sample Square Error}.}
\label{table:ablation_main}
\begin{adjustbox}{max width=0.48\textwidth}
\begin{tabular}{lcccc}
\toprule
\multirow{2}{*}[-0.5\dimexpr \aboverulesep + \belowrulesep + \cmidrulewidth]{\textbf{Method}} & \multicolumn{2}{c}{\textbf{Train}} & \multicolumn{2}{c}{\textbf{Test}} \\
\cmidrule(lr){2-3} \cmidrule(lr){4-5}
\multicolumn{1}{c}{} & \multicolumn{1}{c}{MSE} & \multicolumn{1}{c}{Max SE} & \multicolumn{1}{c}{MSE} & \multicolumn{1}{c}{Max SE} \\
\midrule
\ERM & 0.03 & 0.37 & 0.42 & 11.11 \\
\FL (``$\alpha=\infty$'') & 0.08 & 0.87 & 0.47 & 12.39 \\
\RFL ($\alpha=1$) & 0.05 & 0.66 & 0.44 & 11.14 \\
\RFL ($\alpha=10^{-1}$) & 0.05 & 0.51 & 0.44 & 10.91 \\
\RFL ($\alpha=10^{-2}$) & 0.02 & 0.46 & 0.42 & 11.33 \\
\RFL ($\alpha=10^{-3}$) & \textbf{0.01} & \textbf{0.30} & 0.38 & 11.42 \\
\RFL ($\alpha=10^{-4}$) & 0.06 & 2.58 & \textbf{0.37} & \textbf{10.78} \\
\bottomrule
\end{tabular}
\end{adjustbox}
\end{table}

\vspace{-1ex}
\subsection{How do \FL Solutions Compare to \ERM?}
\label{exps:robustness}

\begin{figure}[h]
    \centering
    \includegraphics[width=\linewidth]{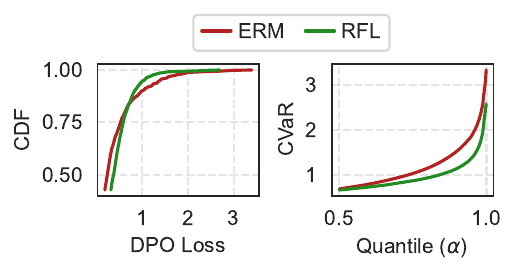}
    \vspace*{-5.5ex}
    \caption{
    Empirical distribution of validation per-sample DPO losses on a fine-tuned Llama3-8B. \textbf{Left}: The empirical Cumulative Density Function (CDF). \textbf{Right}: The empirical Conditional Value at Risk (CVaR).
    \textbf{\FL results in fewer samples with very high losses and a lower average loss for those samples.}
    \captioncomment{The CVaR represents the average loss for samples exceeding each quantile of the loss distribution.}
    }
    \label{fig:cdf_shaping}
\end{figure}

\textbf{Concentrated loss distribution.}
\Cref{fig:cdf_shaping} presents the Cumulative Density Function (CDF) and Conditional Value at Risk (CVaR) for the loss of test samples in the DPO task. For small losses, \ERM's CDF lies above \FL's, indicating that \ERM has a higher proportion of low-loss samples. Conversely, for larger losses, \FL's CDF rises above \ERM's, showing that \FL has fewer samples with very high losses. This suggests that while \ERM performs better on ``easy'' samples, \FL ensures more consistent performance, especially in the tail.

Furthermore, \FL consistently achieves lower CVaR values compared to \ERM, meaning that the average loss for samples with high losses is lower for \FL across all loss percentiles. This highlights \FL's sample-centric nature: outlier samples are less severely impacted than in \ERM. This property makes \FL particularly valuable in applications where consistent performance across all data points is critical.
We observed similar behavior for the training set, and across all tasks (see \Cref{app:experiments}). 

\newpage

\textbf{Multiplier informativity.}
Despite the absence of an objective, which precludes classical sensitivity or shadow price interpretations of Lagrange multipliers, multipliers can still provide insights into the difficulty of satisfying the corresponding constraint.

\begin{figure}[h]
    \centering\includegraphics[width=0.51\linewidth]{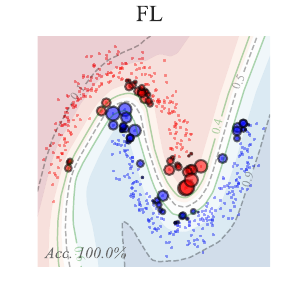}
    \caption{\FL for a two-dimensional classification task. The marker size of each datapoint is proportional to its corresponding multiplier value at the end of training.
    \textbf{Points near the decision boundary have large multipliers, while those farther away have near-zero ones.}
    \captioncomment{The contours correspond to the level curves of the predicted probabilities.}
    }
    \label{fig:two-moons}
\end{figure}

\Cref{fig:two-moons} illustrates the dual variable informativity in a Two-Moons classification task. Samples near the decision boundary, which are harder to classify, have larger multiplier values at the end of training. Hence, similar to support vectors in \SVMs, these samples play a more significant role in shaping the classifier, as their higher multipliers give them greater influence in the primal updates. In contrast, points far from the boundary that are easy to classify have near-zero multipliers and contribute less to the resulting classifier.

\begin{figure}[tbhp]
    \centering
    \begin{minipage}[b]{0.3\linewidth}
        \centering
        \subcaption{$e = 0.15 $, $\lambda = 0.2$}
        \includegraphics[width=\linewidth]{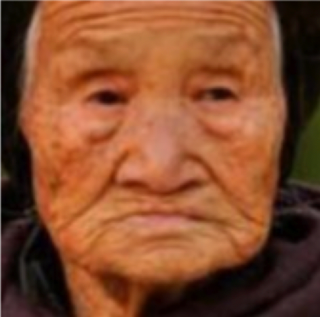}
    \end{minipage}
    \begin{minipage}[b]{0.3\linewidth}
        \centering
        \subcaption{$e = 0.7 $, $\lambda = 0.2$}
        \includegraphics[width=\linewidth]{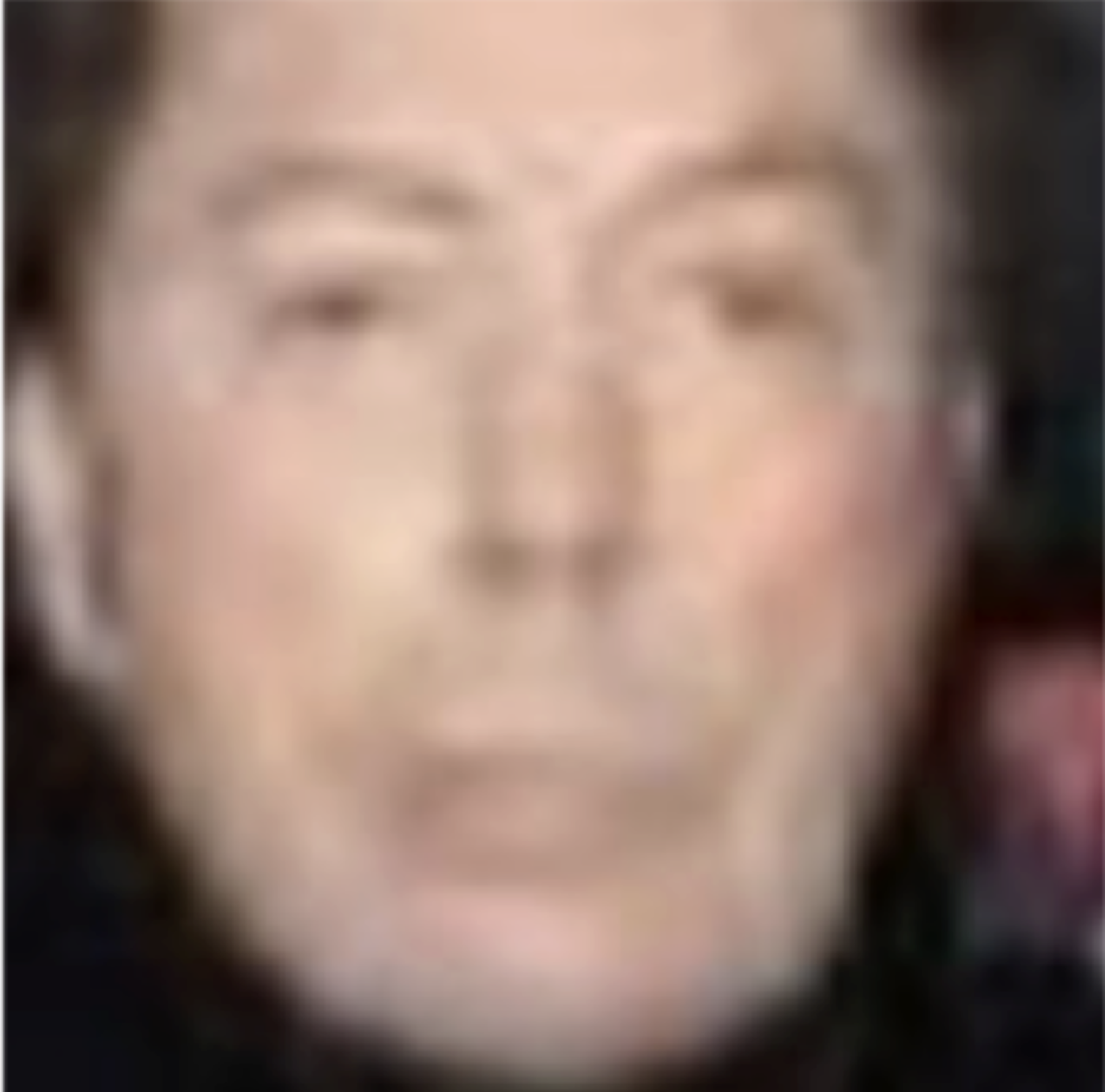}
    \end{minipage}    
    \begin{minipage}[b]{0.3\linewidth}
        \centering
        \subcaption{$e = 0.9$, $\lambda = 6$}
        \includegraphics[width=\linewidth]{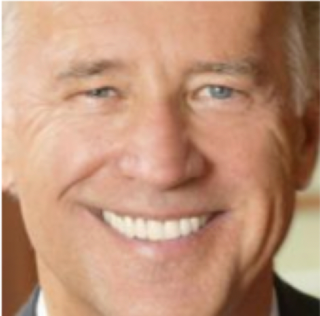}
    \end{minipage}
    \caption{Training samples from UTKFace with large multiplier values at the end of training (top 20), but with low errors. \textbf{These samples are memorized by the model, but are difficult to fit}: (a) a sample with an unusually large age, (b) a blurred image, and (c) a repeated subject yet with different labels. }\label{fig:faces}
    \vspace{-1ex}
\end{figure}

\Cref{fig:faces} shows the loss and multiplier values for some challenging samples in UTKFace age classification. This figure demonstrates that some examples can achieve a small loss yet have a large multiplier—indicating that they were difficult to fit, but the model ultimately managed to do so. The multipliers can help identify outliers or consistently challenging samples that the final loss value alone may not reveal.

Across various tasks, we observe that many samples have zero-valued multipliers towards the end of training, particularly when using \RFL with small $\alpha$ values (see~\Cref{app:experiments}). This implies that the primal updates eventually rely on only a small subset of the data. Beyond simply identifying ``easy'' samples, this observation could be leveraged to improve computational efficiency by pruning these samples from the dataset.

\section{CONCLUSION}

In this work, we introduce Feasible Learning, a novel learning paradigm that frames learning as a constraint satisfaction problem. We show that \FL problems can be solved using a primal-dual approach, which is as computationally efficient as \ERM with gradient descent and offers comparable hyperparameter robustness.

\FL aligns with the growing demand for user-specific performance as machine learning models are increasingly applied in personalized areas like recommender systems and healthcare. Unlike \ERM, \FL directly supports meeting potential regulatory or industry standards that demand a certain level of performance for all users.

We demonstrate that models can learn through \FL, even when using tools originally developed for \ERM, such as modern deep learning architectures and mini-batch optimization techniques. 
Developing algorithmic tools specifically tailored to learning via \FL is an important direction of future research.

\newpage

We show that \FL yields a less heavy-tailed loss distribution than \ERM. We also highlight the informativity of the Lagrange multipliers, as they correlate with the difficulty of fitting each sample.
Other potential benefits of \FL\hspace{-1mm}—which could be explored in future work—may include fairness, due to its sample-centric nature, and calibration, as it does not demand zero training loss.

\subsubsection*{Acknowledgements}

This research was partially supported by an IVADO PhD Excellence Scholarship, the Canada CIFAR AI Chair program (Mila), the NSERC Discovery Grant RGPIN2017-06936, and by Samsung Electronics Co., Ldt. Simon Lacoste-Julien is a CIFAR Associate Fellow in the Learning in Machines \& Brains program.

This research was enabled in part by compute resources, software, and technical help provided by Mila. %

We thank Pedram Khorsandi, Mansi Rankawat, Motahareh Sohrabi, and Rohan Sukumaran for their feedback on the paper.

\subsubsection*{References}

\renewcommand{\bibsection}{}
\bibliographystyle{bib_style}
\bibliography{references}

\section*{Checklist}

\begin{enumerate}
    \item For all models and algorithms presented, check if you include:
    \begin{enumerate}
        \item A clear description of the mathematical setting, assumptions, algorithm, and/or model. \textbf{Yes.} Algorithms: see \S\ref{sec:solving_fl} for \FL and \S\ref{sec:solving_rfl} for \RFL.
        \item An analysis of the properties and complexity (time, space, sample size) of any algorithm. \textbf{Yes.} See \textit{The cost of GDA on \FL} in \S\ref{sec:solving_fl}.
        \item (Optional) Anonymized source code, with specification of all dependencies, including external libraries. \textbf{Yes.} See anonymized source code in the supplementary material.
    \end{enumerate}

 \item For any theoretical claim, check if you include:
 \begin{enumerate}
   \item Statements of the full set of assumptions of all theoretical results. \textbf{Yes.} See \Cref{prop:strong_duality,prop:clamped_quadratic}.
   \item Complete proofs of all theoretical results. \textbf{Yes.} See \Cref{app:proofs}.
   \item Clear explanations of any assumptions. \textbf{Yes.}   
 \end{enumerate}

 \item For all figures and tables that present empirical results, check if you include:
 \begin{enumerate}
   \item The code, data, and instructions needed to reproduce the main experimental results (either in the supplemental material or as a URL). \textbf{Yes.} See the \texttt{scripts} folder in the code.
   \item All the training details (e.g., data splits, hyperparameters, how they were chosen). \textbf{Yes.} See \cref{app:details}.
     \item A clear definition of the specific measure or statistics and error bars (e.g., with respect to the random seed after running experiments multiple times). \textbf{Yes.} See \textit{Experimental uncertainty} in \S\ref{sec:experiments}.
     \item A description of the computing infrastructure used. (e.g., type of GPUs, internal cluster, or cloud provider). \textbf{Yes.} See \textit{Software \& Hardware} in \S\ref{sec:experiments}.
 \end{enumerate}

 \item If you are using existing assets (e.g., code, data, models) or curating/releasing new assets, check if you include:
 \begin{enumerate}
   \item Citations of the creator If your work uses existing assets. \textbf{Yes.} See \textit{Tasks} in \S\ref{sec:experiments}.
   \item The license information of the assets, if applicable. \textbf{Not Applicable.}
   \item New assets either in the supplemental material or as a URL, if applicable. \textbf{Yes.} We include our code.
   \item Information about consent from data providers/curators. \textbf{Not Applicable.}
   \item Discussion of sensible content if applicable, e.g., personally identifiable information or offensive content. \textbf{Not Applicable.}
 \end{enumerate}

 \item If you used crowdsourcing or conducted research with human subjects, check if you include:
 \begin{enumerate}
   \item The full text of instructions given to participants and screenshots. \textbf{Not Applicable.}
   \item Descriptions of potential participant risks, with links to Institutional Review Board (IRB) approvals if applicable. \textbf{Not Applicable.}
   \item The estimated hourly wage paid to participants and the total amount spent on participant compensation. \textbf{Not Applicable.}
 \end{enumerate}

 \end{enumerate}

\appendix

\onecolumn

\addcontentsline{toc}{section}{Appendix}
\part{Appendix} 
\parttoc
\newpage

\section{PROOFS}
\label{app:proofs}

\begin{proof}[Proof of \cref{prop:strong_duality}]
    \label{app:proofs:prop1}
    
    For every $\theta \in \Theta$, $\LagRFL$ is convex in $\vu$ and concave in $\vlambda$. Moreover, we have that $\LagRFL$ is $0$-coercive in $\vu$ for any $\vlambda$, that is, $\LagRFL(\theta,\vu,\vlambda) \rightarrow \infty$ as $\|\vu\| \rightarrow \infty$; and also there exists some fixed $\vu(\theta)$ (big enough depending on $\theta$) such that $\LagRFL(\theta,\vu(\theta),\vlambda) \rightarrow -\infty$ as $\| \vlambda \| \rightarrow \infty$. We can thus apply the existence of a saddle-point theorem from \citepsup[Theorem 4.3.1]{hiriart1996convex} to obtain:
    \begin{equation}
        \min_{\vu \ge \vzero} \,  \max_{\vlambda \ge \vzero} \,\, \LagRFL(\vtheta, \vu, \vlambda) 
        = \max_{\vlambda \ge \vzero} \, \min_{\vu \ge \vzero} \,\, \LagRFL(\vtheta, \vu, \vlambda).
    \end{equation}
    Moreover, the first order optimality condition for the inner minimization $\min_{\vu \ge \vzero} \, \LagRFL(\vtheta, \vu, \vlambda)$ yields $\alpha \vu^* - \vlambda = \vzero$. 
    Since $\vlambda \ge \vzero$, the first order condition is satisfied at the non-negative $\vu^* = \vlambda / \alpha$. It follows that:
    \begin{equation}
        \max_{\vlambda \ge \vzero} \, \min_{\vu \ge \vzero} \,\, \LagRFL(\vtheta, \vu, \vlambda) = \max_{\vlambda \ge \vzero} \LagRFL \left(\vtheta, \frac{1}{\alpha} \vlambda, \vlambda \right) = \max_{\vlambda \ge \vzero} \vlambda\T (\vg(\vtheta) - \vepsilon) - \frac{1}{2 \alpha} ||\vlambda||^2
    \end{equation}
\end{proof}

\begin{proof}[Proof of \cref{prop:clamped_quadratic}]
\label{app:proofs:prop2}

From~\Cref{prop:strong_duality}, it follows that:
\begin{equation}
        \label{eq:clamped_quadratic_erm_proof}
        \min_{\vtheta \in \Theta} \, \min_{\vu \ge \vzero} \, \max_{\vlambda \ge \vzero}\,\,  \LagRFL(\vtheta, \vu, \vlambda) 
        = 
        \min_{\vtheta \in \Theta} \, \max_{\vlambda \ge \vzero}\,\min_{\vu \ge \vzero} \,  \LagRFL(\vtheta, \vu, \vlambda)  =
       \min_{\vtheta \in \Theta} \, \max_{\vlambda \ge \vzero} \, \vlambda\T (\vg(\vtheta) - \vepsilon) - \frac{1}{2 \alpha} ||\vlambda||^2
    \end{equation}
 The first order optimality condition for $\vlambda$ over the non-negative orthant is satisfied at $\vlambda_{\vtheta}^* = \alpha \left[ \vg(\vtheta) - \vepsilon \right]_+$, where $[\, \cdot \,]_+$ is an element-wise projection to $\reals_{\geq 0}$. Substituting this value of $\vlambda$ back in~\cref{eq:clamped_quadratic_erm_proof} yields the desired result.
 
\end{proof}

\section{EXPERIMENTAL DETAILS}
\label{app:details}

Our implementations use PyTorch \citepsup{pytorch} and the Cooper library for constrained optimization \citepsup{gallegoPosada2024cooper}.\footnote{\texttt{\url{https://github.com/cooper-org/cooper}}} Experiments are run on NVIDIA \texttt{L40S} GPUs.
Our code is available at: \texttt{\url{https://github.com/juan43ramirez/feasible-learning}}.

\subsection{Polynomial Regression}
\label{app:polyreg}

We consider the problem of fitting a polynomial $p_{\boldsymbol{a}}(x) = \sum_{j=0}^d a_j x^j_i$ of degree $d$ to a dataset $\{(x_i, y_i)\}_{i=1}^n$ consisting of $n$ samples. The \ERM problem in this setting can be expressed as follows:
\begin{equation}
    \min_{\boldsymbol{a}} \,\, \frac{1}{n} \sum_{i=1}^n \left( p_{\boldsymbol{a}}(x_i) - y_i \right)^2,
\end{equation}
where $\boldsymbol{a}=[a_0, a_1, \dots, a_d]$ represent the coefficients of the polynomial. 

The corresponding \FL problem is:
\begin{equation}
    \label{eq:poly_fl}
    \min_{\boldsymbol{a}} \,\, 0,  \quad \text{s.t. } \left( p_{\boldsymbol{a}}(x_i) - y_i \right)^2 \leq \epsilon, \quad i=1,\dots,n.
\end{equation}

We generate data by sampling $20$ points from a cosine wave and adding Gaussian noise with a standard deviation of $\sigma=0.2$. We fit polynomials of degrees $10$, $20$, and $30$, which represent different model parameterization scenarios: the degree $10$ polynomial is under-parameterized, and the degree $30$ polynomial is over-parameterized.

Due to ill-conditioning, we solve both problems using a standard convex optimization solver: CVXPY's Splitting Conic Solver \citepsup[SCS]{odonoghue:21}.
For \ERM, this solver resorts to a QP solver; for \FL, note that the problem in \cref{eq:poly_fl} constitutes a second-order cone programming problem over $[p_{\boldsymbol{a}}(x_i) - y_i, \epsilon]$ \smash{\citepsup[\S4.4.2]{boyd2004convex}}. 
\FL's constraint value is set to one standard deviation, $\epsilon = \sigma$. 

\subsection{Deep Learning Tasks}

\Cref{tab:expt-details} presents the tasks (datasets and models) considered in our experiments in \S\ref{sec:experiments} and \cref{app:experiments}. We also include the training set size for each dataset, as it corresponds to the number of constraints for each task.

\begin{table}[h!]
\centering
\caption{Datasets considered throughout this work.}
\begin{tabular}{llr}
\toprule
\multicolumn{1}{c}{\textbf{Dataset}} & \multicolumn{1}{c}{\textbf{Model}} & \multicolumn{1}{l}{\textbf{Train Size}} \\
\midrule
CIFAR10~\citepsup{cifar} & ResNet-18~\citepsup{he2016deep} & $50 000$\\
UTKFace~\citepsup{utkface} & ResNet-18~\citepsup{he2016deep} & $4 438$\\
Orca~\citepsup{mukherjee2023orca} & Llama3-8B~\citepsup{dubey2024llama} & $12 859$\\
Two Moons (noise $=0.1$) & MLP (2, 70, 70, 2) & $1 000$\\
\bottomrule
\end{tabular}
\label{tab:expt-details}
\end{table}

\textbf{UTKFace.}
We first perform a 70\%-30\% train-test split, followed by subsampling 25\% of the training set. This is intended to create a simpler task, enabling the ResNet-18 to interpolate most datapoints. However, the resulting dataset still contains some duplicated samples with distinct labels (see \S\ref{exps:dynamics}). Additionally, we normalize the age variable to have zero mean and unit variance.

\textbf{Direct Preference Optimization.}
Since Large Language Models (LLMs) are primarily pre-trained for next token prediction, fine-tuning their weights to align their outputs with human preferences on specific tasks is often desirable. Preferences are typically provided as an input prompt paired with two outputs, one of which is preferred. 

Various techniques exist to increase the likelihood of preferred outputs (see, for example, \citetsup{kaufmann2023survey} and references therein). In particular, \citetsup{rafailov2024dpo} propose an effective supervised approach known as Direct Preference Optimization (DPO). This method aims to maximize the log-likelihood ratio between preferred and dispreferred responses while incorporating a regularization term to penalize deviations from the pre-trained model's outputs. This regularization is standard in preference optimization to mitigate overfitting \smash{\citetsup{ouyang2022training}}, especially given the limited size of preference datasets. The DPO loss is defined as:
\begin{equation*}
\ell_{\text{DPO}} (\bm x, y^+, y^-, \pi_\theta, \pi_{\text{ref}})= \log \sigma\left(\beta \log \frac{\pi_\theta\left(y^+ \mid \bm x\right)}{\pi_{\text {ref }}\left(y^+ \mid \bm x\right)}-\beta \log \frac{\pi_\theta\left(y^- \mid \bm x\right)}{\pi_{\text {ref }}\left(y^- \mid \bm x\right)}\right),
\end{equation*}
where $\bm x$ denotes the input prompt, $y^+$ and $y^-$ represent the preferred and dispreferred completions, and $\pi_\text{ref}$ and $\pi_\theta$ denote the reference (pre-trained) and fine-tuned models. \citetsup{rafailov2024dpo} demonstrate that optimizing this supervised loss is equivalent to optimizing the implicit reward $\hat{r}_\theta(x, y) = \beta \log \frac{\pi_\theta(y \mid x)}{\pi_{\text{ref}}(y \mid x)}$ within a \citetsup{bradley1952rank} preference model.

We fine-tune the 7 billion parameter Llama 3.1 model~\citepsup{dubey2024llama}, and StableLM's Zephyr-3B\footnote{\texttt{\url{https://huggingface.co/stabilityai/stablelm-zephyr-3b}}} on the cleaned version of the Intel Orca dpo pairs dataset.\footnote{\texttt{\url{https://huggingface.co/datasets/argilla/distilabel-intel-orca-dpo-pairs}}} This synthetic preference dataset comprises 6k prompts across various domains and tasks, along with their corresponding outputs from ChatGPT and Llama2-13B. In this version of the dataset, ChatGPT is used to score outputs and the preferred choices are designated based on these scores.  
Because preference datasets are often small, a KL regularization that penalizes deviations from the pre-trained model's outputs is used to mitigate overfitting. In our experiments, the regularization coefficient $\beta$ was set to $0.1$. We use Huggingface Transformer Reinforcement Learning
(trl) library.\footnote{\texttt{\url{https://github.com/huggingface/trl}}}

To reduce hardware requirements for fine-tuning, we apply Low-Rank Adaptation (LoRA), a popular parameter-efficient fine-tuning approach that utilizes a low-rank parametrization of weight matrix updates \citepsup{hu2021lora}. This method decreases the number of learnable parameters, eliminating the need for gradient computation and optimizer state maintenance for most parameters. We further reduce resource requirements by quantizing the pre-trained model to four-bit precision, as proposed by \citetsup{dettmers2024qlora} and implemented in Hugging Face's Parameter-Efficient Fine-Tuning library.\footnote{\texttt{\url{https://github.com/huggingface/peft}}}

\textbf{Hyper-parameter choices.} 
\Cref{tab:primal-hyperparams} lists the primal hyperparameters used to train our models. 
We employ a cosine learning rate scheduler for CIFAR10 classification and Orca DPO experiments. 
\Cref{tab:expt-hyperparams} details the dual hyperparameters employed in training our \FL and \RFL models. 
Additional hyperparameters for our DPO experiments are provided in \Cref{a:tab:hparams:dpo}.

\begin{table}[t]
\vspace{-1ex}
\centering
\caption{Primal optimization hyper-parameters. \captioncomment{To address numerical issues, we use a step size of $\lrp / 10$ for the \CSERM experiments on the UTKFace and CIFAR-10 datasets, where $\lrp$ denotes the reported primal learning rate.}}
\vspace{-1ex}
\begin{tabular}{llllll}
\toprule
\textbf{Dataset} & \textbf{Epochs} & \textbf{Batch Size} & \textbf{Optimizer} & \textbf{Step-size} & \textbf{Weight decay} \\
\midrule
CIFAR10 & $200$ & $128$ & SGD & $1\cdot10^{-1}$ & $5\cdot10^{-4}$ \\ 
UTKFace & $150$ & $128$ & AdamW & $1\cdot10^{-4}$ & $0$ \\ 
Orca & $20$ & $16$ & AdamW & $5\cdot10^{-6}$ & $0$ \\ 
Two Moons & $250$ & $512$ & AdamW & $5\cdot10^{-4}$ & $0$ \\ 
\bottomrule
\end{tabular}
\label{tab:primal-hyperparams}
\end{table}

\begin{table}[t]
\centering
\caption{Dual optimization hyper-parameters for \FL and \RFL experiments.}
\vspace{-1ex}
\begin{tabular}{lcl}
\toprule
\textbf{Dataset} & \textbf{Optimizer}  & \textbf{Step-Size}\\
\midrule
CIFAR-10 & SGD & $1\cdot10^{-4}$\\
UTKFace & SGD & $1\cdot10^{-3}$\\
Orca & SGD & $1\cdot10^{-1}$ \\
Two Moons & SGD & $1\cdot10^{-2}$\\
\bottomrule
\end{tabular}
\label{tab:expt-hyperparams}
\end{table}

\begin{table}[t!]
\centering
\caption{Hyperparameter configurations for Orca DPO pairs.}
\vspace{-1ex}
\begin{tabular}{ccccccc}
\toprule
\textbf{LoRA} $\alpha$ & \textbf{LoRA rank} & \textbf{Scheduler} & \textbf{Warmup Steps} & \textbf{DPO} $\beta$ \\
\midrule
$1$                    & $8$                & Cosine             & $200$                 & $0.1$ \\   
\bottomrule
\end{tabular}
\label{a:tab:hparams:dpo}
\end{table}

\section{ADDITIONAL EXPERIMENTS}
\label{app:experiments}

This section complements \S\ref{sec:experiments}. For each subsection—\cref{app:cifar} for CIFAR10 classification, \cref{app:utk} for UTKFace age regression, and \cref{app:llms} for direct preference optimization of large language models—we address questions \textbf{Q1}-\textbf{Q3} from \S\ref{sec:intro}. The results presented here support the findings from \S\ref{sec:experiments}, specifically: \blobletter{1}: models trained with \FL and \RFL achieve comparable train and test performance to \ERM, \blobletter{2}: \RFL (with appropriate $\alpha$ choices) can succeed where \FL fails, and \blobletter{3}: Feasible Learning can help shape the distribution of losses, generally producing a less heavy-tailed distribution.

Moreover, we include \Cref{fig:cifar_robustness} for CIFAR10 and \Cref{fig:utk_robustness} for UTKFace, which show the training losses recovered by \ERM, \FL, \RFL, and \CSERM for different choices of the primal and dual learning rates. These figures demonstrate that \FL and \RFL are similarly robust to the choice of primal learning rate as \ERM, while also being fairly robust to the choice of dual learning rate.

\vspace{-1ex}
\subsection{CIFAR10}
\label{app:cifar}

\textbf{Can we Learn with Feasible Learning?}
\Cref{table:CIFAR10} presents the performance at the end of training for the CIFAR10 classification task. We report the outcomes of \ERM, as well as \FL, \RFL($\alpha=1$), and \CSERM under two constraint levels: $\epsilon=0.51$, which requires the model to assign a probability of at least 0.6 to the correct class, and $\epsilon=0$, where the model is required to assign a probability of 1 to the correct label, aligning with \ERM's solution set if data interpolation is possible.

\begin{table}[t!]
\centering
\caption{Final performance for CIFAR10 experiments. \textbf{\FL and \RFL achieve comparable average losses and accuracies to \ERM, on both the training and test sets.} \ERM, \FL, and \RFL interpolate the training data in this task, achieving perfect training accuracy and a nearly zero training loss on all samples. \captioncomment{This is an extended version of \Cref{table:main} in \S\ref{exps:learning}.} }
\label{table:CIFAR10}
\vspace{-1ex}
\begin{adjustbox}{max width=\textwidth}
\begin{tabular}{lcrrrrrr}
\toprule
\multirow{2}{*}[-0.5\dimexpr \aboverulesep + \belowrulesep + \cmidrulewidth]{\textbf{Method}} & \multirow{2}{*}[-0.5\dimexpr \aboverulesep + \belowrulesep + \cmidrulewidth]{\textbf{$\epsilon$}} & \multicolumn{3}{c}{\textbf{Train}} & \multicolumn{3}{c}{\textbf{Test}} \\
\cmidrule(lr){3-5} \cmidrule(lr){6-8}
\multicolumn{1}{c}{} & \multicolumn{1}{c}{} & \multicolumn{1}{c}{Avg. CE} & \multicolumn{1}{c}{Max CE} & \multicolumn{1}{c}{Acc.} & \multicolumn{1}{c}{Avg. CE} & \multicolumn{1}{c}{Max CE} & \multicolumn{1}{c}{Acc.} \\
\midrule
\ERM &  & 0.002 {\color{gray} $\pm$ 0.000} & 2.369 {\color{gray} $\pm$ 0.647} & 1.000 {\color{gray} $\pm$ 0.000} & 0.298 {\color{gray} $\pm$ 0.009} & 12.830 {\color{gray} $\pm$ 0.443} & 0.932 {\color{gray} $\pm$ 0.002} \\
\midrule
\FL & $0.00$ & 0.003 {\color{gray} $\pm$ 0.000} & 3.084 {\color{gray} $\pm$ 1.064} & 1.000 {\color{gray} $\pm$ 0.000} & 0.331 {\color{gray} $\pm$ 0.006} & 15.533 {\color{gray} $\pm$ 0.642} & 0.927 {\color{gray} $\pm$ 0.001} \\
\RFL ($\alpha = 1$) & $0.00$ & 0.003 {\color{gray} $\pm$ 0.000} & 3.224 {\color{gray} $\pm$ 1.311} & 1.000 {\color{gray} $\pm$ 0.000} & 0.337 {\color{gray} $\pm$ 0.012} & 16.235 {\color{gray} $\pm$ 1.510} & 0.927 {\color{gray} $\pm$ 0.002} \\
\CSERM & $0.00$ & 0.065 {\color{gray} $\pm$ 0.091} & 2.667 {\color{gray} $\pm$ 0.926} & 0.988 {\color{gray} $\pm$ 0.024} & 0.422 {\color{gray} $\pm$ 0.136} & 12.092 {\color{gray} $\pm$ 0.628} & 0.870 {\color{gray} $\pm$ 0.048} \\
\midrule
\FL & $0.51$ & 0.015 {\color{gray} $\pm$ 0.001} & 4.991 {\color{gray} $\pm$ 0.814} & 0.997 {\color{gray} $\pm$ 0.000} & 0.342 {\color{gray} $\pm$ 0.007} & 14.874 {\color{gray} $\pm$ 1.541} & 0.918 {\color{gray} $\pm$ 0.001} \\
\RFL ($\alpha = 1$) & $0.51$ & 0.014 {\color{gray} $\pm$ 0.000} & 5.713 {\color{gray} $\pm$ 0.946} & 0.998 {\color{gray} $\pm$ 0.000} & 0.351 {\color{gray} $\pm$ 0.009} & 14.139 {\color{gray} $\pm$ 0.735} & 0.917 {\color{gray} $\pm$ 0.001} \\
\CSERM & $0.51$ & 0.337 {\color{gray} $\pm$ 0.018} & 1.724 {\color{gray} $\pm$ 0.366} & 0.978 {\color{gray} $\pm$ 0.010} & 0.522 {\color{gray} $\pm$ 0.018} & 8.770 {\color{gray} $\pm$ 0.851} & 0.881 {\color{gray} $\pm$ 0.014} \\
\bottomrule
\end{tabular}
\end{adjustbox}
\vspace{-1ex}
\end{table}

We observe that \ERM generates models that nearly interpolate the training data, achieving near-zero loss on most samples. However, the maximum loss indicates that some training points are not correctly classified by \ERM. Note that these misclassifications are not evident in the accuracy column, as we report accuracy only up to three significant digits for a training dataset of $60 000$ samples.

\begin{figure}[t!]
    \vspace{-1ex}
    \centering
    \includegraphics{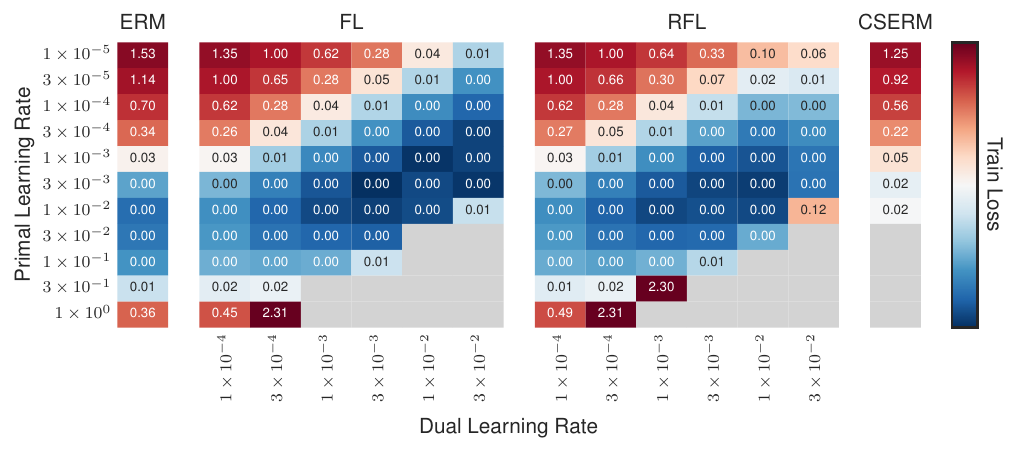}
    \caption{Average training losses at the end of training on the CIFAR10 classification task. \textbf{Similar to \ERM, both \FL and \RFL achieve good training losses across a wide range of orders of magnitude for the primal learning rate.} Furthermore, \FL and \RFL demonstrate good performance for multiple choices of the dual learning rate. \captioncomment{A grey square indicates a configuration that diverged. $\epsilon=0.51$.}}
    \label{fig:cifar_robustness}
\end{figure}

At $\epsilon=0$, \FL and \RFL achieve near-perfect training accuracy and similar average and maximum train losses compared to \ERM. However, their test performance is slightly worse on all metrics. Moreover, \CSERM yields significantly worse training and test performance, which we attribute to potential numerical instabilities caused by squaring the loss functions (leading to exploding or vanishing gradients). \Cref{fig:cifar_robustness}, presented later, highlights the difficulties in properly tuning the primal learning rate for \CSERM.

At $\epsilon=0.51$, the training performance of \FL and \RFL remains comparable to that of \ERM; however, we observe greater degradation compared to $\epsilon=0$. This trend extends to the test set as well. These results suggest that the constraint was too loose. A tighter, though not necessarily zero, constraint level could potentially enhance both training and test performance.

\textbf{How does Resilience Help?}
\Cref{table:ablation_CIFAR10_0.6,table:ablation_CIFAR10_1} present an ablation study of \RFL's $\alpha$ value for the CIFAR10 classification task, with $\epsilon=0$ and $\epsilon=0.51$, respectively.

\begin{table}[t!]
\centering
\caption{Final performance for CIFAR10 classification. \captioncomment{$\epsilon=0$.}}
\label{table:ablation_CIFAR10_1}
\vspace{-1ex}
\begin{adjustbox}{max width=\textwidth}
\begin{tabular}{lrrrrrr}
\toprule
\multirow{2}{*}[-0.5\dimexpr \aboverulesep + \belowrulesep + \cmidrulewidth]{\textbf{Method}} & \multicolumn{3}{c}{\textbf{Train}} & \multicolumn{3}{c}{\textbf{Test}} \\
\cmidrule(lr){2-4} \cmidrule(lr){5-7}
\multicolumn{1}{c}{} & \multicolumn{1}{c}{Avg. CE} & \multicolumn{1}{c}{Max CE} & \multicolumn{1}{c}{Acc.} & \multicolumn{1}{c}{Avg. CE} & \multicolumn{1}{c}{Max CE} & \multicolumn{1}{c}{Acc.} \\
\midrule
\ERM & 0.002 {\color{gray} $\pm$ 0.000} & 2.369 {\color{gray} $\pm$ 0.579} & 1.000 {\color{gray} $\pm$ 0.000} & 0.298 {\color{gray} $\pm$ 0.008} & 12.830 {\color{gray} $\pm$ 0.396} & 0.932 {\color{gray} $\pm$ 0.002} \\
\FL (``$\alpha\hspace{-0.5ex}=\hspace{-0.5ex}\infty$'') & 0.003 {\color{gray} $\pm$ 0.000} & 3.428 {\color{gray} $\pm$ 1.086} & 1.000 {\color{gray} $\pm$ 0.000} & 0.332 {\color{gray} $\pm$ 0.007} & 16.091 {\color{gray} $\pm$ 1.177} & 0.927 {\color{gray} $\pm$ 0.002} \\
\RFL ($\alpha\hspace{-0.5ex}=\hspace{-0.5ex}1$) & 0.003 {\color{gray} $\pm$ 0.000} & 3.673 {\color{gray} $\pm$ 1.759} & 1.000 {\color{gray} $\pm$ 0.000} & 0.322 {\color{gray} $\pm$ 0.008} & 15.013 {\color{gray} $\pm$ 0.907} & 0.927 {\color{gray} $\pm$ 0.002} \\
\RFL ($\alpha\hspace{-0.5ex}=\hspace{-0.5ex}10^{-1}$) & 0.003 {\color{gray} $\pm$ 0.000} & 3.157 {\color{gray} $\pm$ 0.839} & 1.000 {\color{gray} $\pm$ 0.000} & 0.326 {\color{gray} $\pm$ 0.007} & 14.972 {\color{gray} $\pm$ 0.740} & 0.928 {\color{gray} $\pm$ 0.001} \\
\RFL ($\alpha\hspace{-0.5ex}=\hspace{-0.5ex}10^{-2}$) & 0.005 {\color{gray} $\pm$ 0.000} & 3.022 {\color{gray} $\pm$ 0.787} & 1.000 {\color{gray} $\pm$ 0.000} & 0.306 {\color{gray} $\pm$ 0.007} & 12.502 {\color{gray} $\pm$ 1.120} & 0.928 {\color{gray} $\pm$ 0.002} \\
\RFL ($\alpha\hspace{-0.5ex}=\hspace{-0.5ex}10^{-3}$) & 0.078 {\color{gray} $\pm$ 0.000} & 2.695 {\color{gray} $\pm$ 0.421} & 1.000 {\color{gray} $\pm$ 0.000} & 0.309 {\color{gray} $\pm$ 0.009} & 7.168 {\color{gray} $\pm$ 0.236} & 0.928 {\color{gray} $\pm$ 0.002} \\
\bottomrule
\end{tabular}
\end{adjustbox}
\vspace{-1ex}
\end{table}

\begin{table}[t!]
\centering
\caption{Final performance for CIFAR10 classification. \captioncomment{$\epsilon=0.51$.}}
\vspace{-1ex}
\label{table:ablation_CIFAR10_0.6}
\begin{adjustbox}{max width=\textwidth}
\begin{tabular}{lrrrrrr}
\toprule
\multirow{2}{*}[-0.5\dimexpr \aboverulesep + \belowrulesep + \cmidrulewidth]{\textbf{Method}} & \multicolumn{3}{c}{\textbf{Train}} & \multicolumn{3}{c}{\textbf{Test}} \\
\cmidrule(lr){2-4} \cmidrule(lr){5-7}
\multicolumn{1}{c}{} & \multicolumn{1}{c}{Avg. CE} & \multicolumn{1}{c}{Max CE} & \multicolumn{1}{c}{Acc.} & \multicolumn{1}{c}{Avg. CE} & \multicolumn{1}{c}{Max CE} & \multicolumn{1}{c}{Acc.} \\
\midrule
\ERM & 0.002 {\color{gray} $\pm$ 0.000} & 2.369 {\color{gray} $\pm$ 0.579} & 1.000 {\color{gray} $\pm$ 0.000} & 0.298 {\color{gray} $\pm$ 0.008} & 12.830 {\color{gray} $\pm$ 0.396} & 0.932 {\color{gray} $\pm$ 0.002} \\
\FL (``$\alpha\hspace{-0.5ex}=\hspace{-0.5ex}\infty$'') & 0.014 {\color{gray} $\pm$ 0.001} & 5.422 {\color{gray} $\pm$ 0.565} & 0.998 {\color{gray} $\pm$ 0.000} & 0.341 {\color{gray} $\pm$ 0.007} & 14.671 {\color{gray} $\pm$ 1.364} & 0.918 {\color{gray} $\pm$ 0.003} \\
\RFL ($\alpha\hspace{-0.5ex}=\hspace{-0.5ex}1$) & 0.014 {\color{gray} $\pm$ 0.001} & 5.401 {\color{gray} $\pm$ 1.052} & 0.998 {\color{gray} $\pm$ 0.000} & 0.345 {\color{gray} $\pm$ 0.007} & 13.720 {\color{gray} $\pm$ 0.604} & 0.917 {\color{gray} $\pm$ 0.001} \\
\RFL ($\alpha\hspace{-0.5ex}=\hspace{-0.5ex}10^{-1}$) & 0.016 {\color{gray} $\pm$ 0.000} & 4.763 {\color{gray} $\pm$ 0.049} & 0.998 {\color{gray} $\pm$ 0.000} & 0.346 {\color{gray} $\pm$ 0.006} & 13.393 {\color{gray} $\pm$ 0.719} & 0.916 {\color{gray} $\pm$ 0.001} \\
\RFL ($\alpha\hspace{-0.5ex}=\hspace{-0.5ex}10^{-2}$) & 0.064 {\color{gray} $\pm$ 0.004} & 4.543 {\color{gray} $\pm$ 0.323} & 0.997 {\color{gray} $\pm$ 0.000} & 0.380 {\color{gray} $\pm$ 0.006} & 10.734 {\color{gray} $\pm$ 0.539} & 0.914 {\color{gray} $\pm$ 0.003} \\
\RFL ($\alpha\hspace{-0.5ex}=\hspace{-0.5ex}10^{-3}$) & 0.479 {\color{gray} $\pm$ 0.002} & 2.013 {\color{gray} $\pm$ 0.127} & 0.999 {\color{gray} $\pm$ 0.000} & 0.646 {\color{gray} $\pm$ 0.004} & 5.293 {\color{gray} $\pm$ 0.238} & 0.925 {\color{gray} $\pm$ 0.002} \\
\bottomrule
\end{tabular}
\end{adjustbox}
\vspace{-1ex}
\end{table}

At $\epsilon=0$, all values of $\alpha$ result in models with near-perfect training accuracy. Smaller $\alpha$ values lead to lower maximum losses, indicating a more compact loss distribution, but with increased average loss. Decreasing $\alpha$ also improves both average and maximum test losses.

From the \RFL problem's perspective, different $\alpha$ values do not change the solutions. However, algorithmically, smaller $\alpha$ values lead to more multipliers going to zero, which reduces the pressure to overfit the training data. This reduction in pressure results in a higher average loss, as fewer points are interpolated. However, it also reduces the maximum losses, as overfitting the majority of the data can lead to high-confidence misclassifications of points in the tail, particularly if they are noisy or mislabeled.

At $\epsilon=0.51$, we observe a slight degradation in both training and test performance compared to \FL and \RFL at $\epsilon=0$. Since the model can effectively interpolate this dataset, we hypothesize that the constraint level of $\epsilon=0.51$ is too loose to achieve optimal performance. However, to account for potentially misclassified hard-to-fit samples, we hypothesize that some $\epsilon>0$ may yield better performance than both $\epsilon=0.51$ and $\epsilon=0$.

\begin{figure}[t!]
\vspace{-1ex}
\centering
\begin{subfigure}{.5\textwidth}
  \centering
  \includegraphics{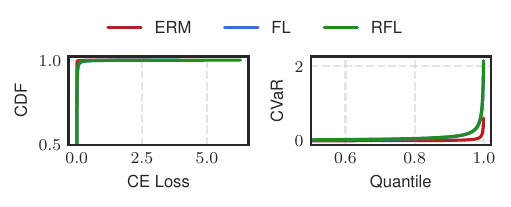}
  \caption{(a) Train CDF and CVaR.}
  \label{fig:cifar_cdf_train}
\end{subfigure}%
\begin{subfigure}{.5\textwidth}
  \centering
  \includegraphics{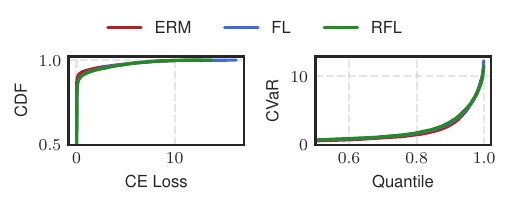}
  \caption{(b) Test CDF and CVaR.}
  \label{fig:cifar_cdf_test}
\end{subfigure}
\caption{Empirical distribution of per-sample cross-entropy (CE) losses on a CIFAR10 classification task. \textbf{Left}: The empirical Cumulative Density Function (CDF). \textbf{Right}: The empirical Conditional Value at Risk (CVaR). \captioncomment{The CVaR represents the average loss for samples exceeding each quantile of the loss distribution. $\epsilon=0.51$. \RFL's $\alpha = 1$.}}
\label{fig:cifar_cdf}
\vspace{-1ex}

\end{figure}

\textbf{How do \FL Solutions Compare to \ERM?}
\Cref{fig:cifar_cdf} displays the CDFs and CVaRs for the train and test sets in this task, including \ERM, \FL, and \RFL($\alpha=1$). In training, we observe slightly better CDF and CVaR curves for \ERM compared to \FL and \RFL. This aligns with the results in \Cref{table:CIFAR10}, indicating that \ERM outperforms these methods in both average and maximum training losses. A similar trend is observed in the test set.

\textbf{Hyper-parameter Robustness.}
\Cref{fig:cifar_robustness} shows the training losses at the end of training on the CIFAR10 classification task (with $\epsilon=0.51$), illustrating the sensitivity of \FL and \RFL to the choice of primal and dual learning rates. Similar to \ERM, both \FL and \RFL achieve good training losses across a wide range of orders of magnitude for the primal learning rate. Furthermore, \FL and \RFL demonstrate good performance for multiple choices of the dual learning rate. 
This indicates that \FL and \RFL are fairly robust to these hyper-parameters, thus contributing to their applicability in practice. 

We also observe that large primal learning rate choices cause divergence in \CSERM runs. We attribute this to the squaring of the loss which can lead to numerical issues for large losses. Additionally, for values that do not diverge, we see degraded performance compared to other approaches. While the squared loss amplifies large losses, it also results in vanishing gradients for small losses, making it harder to overfit the training samples. This situation highlights the practical advantages of \RFL over \CSERM, despite them being equivalent problem formulations.

\subsection{UTKFace}
\label{app:utk}

\textbf{Can we Learn with Feasible Learning?}
\Cref{table:UTKFace} shows the performance at the end of training for the UTKFace regression task. We observe that both \FL and \RFL outperform \ERM in terms of average loss on the training and test sets, while maintaining comparable maximum losses to \ERM. Additionally, we highlight that \RFL achieves this performance with low variance in metrics across runs, indicating more robust dynamics than \FL.

\begin{table}[t!]
\centering
\caption{
Final performance for UTKFace experiments. \textbf{\FL and \RFL outperform \ERM in terms of MSE, on both the training and test sets.}
}
\label{table:UTKFace}
\begin{adjustbox}{max width=\textwidth}
\begin{tabular}{lcrrrr}
\toprule
\multirow{2}{*}[-0.5\dimexpr \aboverulesep + \belowrulesep + \cmidrulewidth]{\textbf{Method}} & \multirow{2}{*}[-0.5\dimexpr \aboverulesep + \belowrulesep + \cmidrulewidth]{\textbf{$\epsilon$}} & \multicolumn{2}{c}{\textbf{Train}} & \multicolumn{2}{c}{\textbf{Test}} \\
\cmidrule(lr){3-4} \cmidrule(lr){5-6}
\multicolumn{1}{c}{} & \multicolumn{1}{c}{} & \multicolumn{1}{c}{Avg. SE} & \multicolumn{1}{c}{Max SE} & \multicolumn{1}{c}{Avg. SE} & \multicolumn{1}{c}{Max SE} \\
\midrule
\ERM &  & 0.060 {\color{gray} $\pm$ 0.017} & 0.438 {\color{gray} $\pm$ 0.021} & 0.449 {\color{gray} $\pm$ 0.016} & 10.335 {\color{gray} $\pm$ 0.398} \\
\midrule
\FL & $0.00$ & 0.026 {\color{gray} $\pm$ 0.041} & 0.630 {\color{gray} $\pm$ 0.190} & 0.407 {\color{gray} $\pm$ 0.048} & 11.361 {\color{gray} $\pm$ 0.719} \\
\RFL ($\alpha=10^{-3}$) & $0.00$ & 0.006 {\color{gray} $\pm$ 0.002} & 0.299 {\color{gray} $\pm$ 0.022} & 0.379 {\color{gray} $\pm$ 0.007} & 11.419 {\color{gray} $\pm$ 0.589} \\
\CSERM & $0.00$ & 0.005 {\color{gray} $\pm$ 0.000} & 0.343 {\color{gray} $\pm$ 0.019} & 0.565 {\color{gray} $\pm$ 0.000} & 15.341 {\color{gray} $\pm$ 0.489} \\
\midrule
\FL & $0.02$ & 0.025 {\color{gray} $\pm$ 0.017} & 0.863 {\color{gray} $\pm$ 0.367} & 0.396 {\color{gray} $\pm$ 0.004} & 13.359 {\color{gray} $\pm$ 2.010} \\
\RFL ($\alpha=10^{-1}$) & $0.02$ & 0.011 {\color{gray} $\pm$ 0.002} & 0.526 {\color{gray} $\pm$ 0.031} & 0.409 {\color{gray} $\pm$ 0.002} & 10.473 {\color{gray} $\pm$ 0.877} \\
\CSERM & $0.02$ & 0.008 {\color{gray} $\pm$ 0.001} & 0.344 {\color{gray} $\pm$ 0.053} & 0.572 {\color{gray} $\pm$ 0.004} & 15.132 {\color{gray} $\pm$ 0.489} \\
\bottomrule
\end{tabular}
\end{adjustbox}
\end{table}

\textbf{How does Resilience Help?}
\Cref{table:ablation_UTKFace_0,table:ablation_UTKFace_0.02}  present an ablation on \RFL's $\alpha$ value for the UTKFace regression task, with $\epsilon=0$ and $\epsilon=0.02$, respectively.

At $\epsilon=0$, we observe that \RFL consistently outperforms \FL on both the training and test sets. The best training performance occurs at $\alpha=10^{-3}$, while the best test performance is achieved at $\alpha=10^{-4}$. 
We attribute \RFL's advantage over \FL to the problem's infeasibility at $\epsilon=0$ (the dataset contains duplicated samples with different labels) which leads to poor optimization dynamics in \FL.

At $\epsilon=0.02$, we observe that \RFL does not necessarily outperform \FL. In this scenario, where feasible solutions may exist, \FL is capable of learning well-performing solutions. However, with appropriate choices of $\alpha$, \RFL can still outperform \FL on both the training and test sets (e.g., $\alpha=10^{-1}$ for training and $\alpha=10^{-4}$ for testing). That said, finding these well-performing solutions requires tuning the additional hyper-parameter $\alpha$.

\begin{table}[t!]
\centering
\caption{Final performance for UTKFace regression. \captioncomment{This is an extended version of \Cref{table:ablation_main} in \S\ref{exps:dynamics}. $\epsilon=0$.}}
\label{table:ablation_UTKFace_0}
\begin{adjustbox}{max width=\textwidth}
\begin{tabular}{lrrrr}
\toprule
\multirow{2}{*}[-0.5\dimexpr \aboverulesep + \belowrulesep + \cmidrulewidth]{\textbf{Method}} & \multicolumn{2}{c}{\textbf{Train}} & \multicolumn{2}{c}{\textbf{Test}} \\
\cmidrule(lr){2-3} \cmidrule(lr){4-5}
\multicolumn{1}{c}{} & \multicolumn{1}{c}{Avg. SE} & \multicolumn{1}{c}{Max SE} & \multicolumn{1}{c}{Avg. SE} & \multicolumn{1}{c}{Max SE} \\
\midrule
\ERM & 0.026 {\color{gray} $\pm$ 0.004} & 0.368 {\color{gray} $\pm$ 0.049} & 0.417 {\color{gray} $\pm$ 0.004} & 11.115 {\color{gray} $\pm$ 1.231} \\
\FL (``$\alpha=\infty$'') & 0.083 {\color{gray} $\pm$ 0.034} & 0.868 {\color{gray} $\pm$ 0.208} & 0.474 {\color{gray} $\pm$ 0.041} & 12.393 {\color{gray} $\pm$ 0.520} \\
\RFL ($\alpha=1$) & 0.050 {\color{gray} $\pm$ 0.017} & 0.658 {\color{gray} $\pm$ 0.321} & 0.442 {\color{gray} $\pm$ 0.022} & 11.139 {\color{gray} $\pm$ 0.841} \\
\RFL ($\alpha=10^{-1}$) & 0.048 {\color{gray} $\pm$ 0.013} & 0.511 {\color{gray} $\pm$ 0.038} & 0.444 {\color{gray} $\pm$ 0.017} & 10.912 {\color{gray} $\pm$ 0.838} \\
\RFL ($\alpha=10^{-2}$) & 0.017 {\color{gray} $\pm$ 0.014} & 0.459 {\color{gray} $\pm$ 0.048} & 0.420 {\color{gray} $\pm$ 0.030} & 11.327 {\color{gray} $\pm$ 1.275} \\
\RFL ($\alpha=10^{-3}$) & 0.006 {\color{gray} $\pm$ 0.002} & 0.299 {\color{gray} $\pm$ 0.020} & 0.379 {\color{gray} $\pm$ 0.007} & 11.419 {\color{gray} $\pm$ 0.538} \\
\RFL ($\alpha=10^{-4}$) & 0.064 {\color{gray} $\pm$ 0.079} & 2.584 {\color{gray} $\pm$ 3.253} & 0.368 {\color{gray} $\pm$ 0.012} & 10.780 {\color{gray} $\pm$ 0.933} \\
\bottomrule
\end{tabular}
\end{adjustbox}
\end{table}

\begin{table}[t!]
\centering
\caption{Final performance for UTKFace regression. \captioncomment{$\epsilon=0.02$.}}
\label{table:ablation_UTKFace_0.02}
\begin{adjustbox}{max width=\textwidth}
\begin{tabular}{lrrrr}
\toprule
\multirow{2}{*}[-0.5\dimexpr \aboverulesep + \belowrulesep + \cmidrulewidth]{\textbf{Method}} & \multicolumn{2}{c}{\textbf{Train}} & \multicolumn{2}{c}{\textbf{Test}} \\
\cmidrule(lr){2-3} \cmidrule(lr){4-5}
\multicolumn{1}{c}{} & \multicolumn{1}{c}{Avg. SE} & \multicolumn{1}{c}{Max SE} & \multicolumn{1}{c}{Avg. SE} & \multicolumn{1}{c}{Max SE} \\
\midrule
\ERM & 0.026 {\color{gray} $\pm$ 0.004} & 0.368 {\color{gray} $\pm$ 0.049} & 0.417 {\color{gray} $\pm$ 0.004} & 11.115 {\color{gray} $\pm$ 1.231} \\
\FL (``$\alpha=\infty$'') & 0.042 {\color{gray} $\pm$ 0.004} & 1.255 {\color{gray} $\pm$ 0.019} & 0.399 {\color{gray} $\pm$ 0.003} & 15.080 {\color{gray} $\pm$ 0.984} \\
\RFL ($\alpha=1$) & 0.017 {\color{gray} $\pm$ 0.002} & 0.443 {\color{gray} $\pm$ 0.045} & 0.387 {\color{gray} $\pm$ 0.007} & 9.667 {\color{gray} $\pm$ 1.000} \\
\RFL ($\alpha=10^{-1}$) & 0.011 {\color{gray} $\pm$ 0.002} & 0.526 {\color{gray} $\pm$ 0.031} & 0.409 {\color{gray} $\pm$ 0.002} & 10.473 {\color{gray} $\pm$ 0.877} \\
\RFL ($\alpha=10^{-2}$) & 0.023 {\color{gray} $\pm$ 0.017} & 0.376 {\color{gray} $\pm$ 0.031} & 0.433 {\color{gray} $\pm$ 0.030} & 10.695 {\color{gray} $\pm$ 0.724} \\
\RFL ($\alpha=10^{-3}$) & 0.054 {\color{gray} $\pm$ 0.088} & 1.524 {\color{gray} $\pm$ 2.500} & 0.392 {\color{gray} $\pm$ 0.023} & 12.237 {\color{gray} $\pm$ 1.577} \\
\RFL ($\alpha=10^{-4}$) & 0.078 {\color{gray} $\pm$ 0.089} & 1.382 {\color{gray} $\pm$ 2.047} & 0.381 {\color{gray} $\pm$ 0.038} & 11.213 {\color{gray} $\pm$ 1.674} \\
\bottomrule
\end{tabular}
\end{adjustbox}
\end{table}

\begin{figure}[t!]
\centering
\begin{subfigure}{.5\textwidth}
  \centering
  \includegraphics{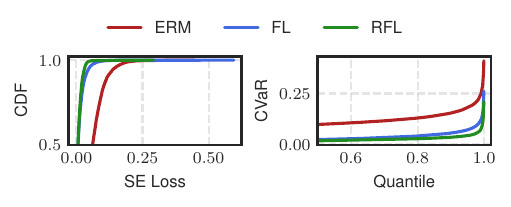}
  \caption{(a) Train CDF and CVaR.}
  \label{fig:utk_cdf_train}
\end{subfigure}%
\begin{subfigure}{.5\textwidth}
  \centering
  \includegraphics{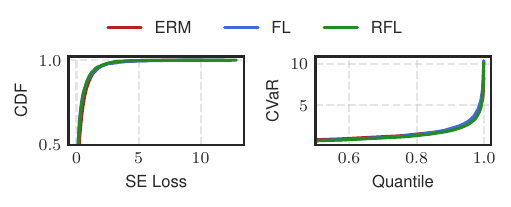}
  \caption{(b) Test CDF and CVaR.}
  \label{fig:utk_cdf_test}
\end{subfigure}
\caption{Empirical distribution of per-sample squared error (SE) losses on a UTKFace age regression task. \textbf{Left}: The empirical Cumulative Density Function (CDF). \textbf{Right}: The empirical Conditional Value at Risk (CVaR). \captioncomment{The CVaR represents the average loss for samples exceeding each quantile of the loss distribution. $\epsilon=0.02$. \RFL's $\alpha = 10^{-1}$.}}
\label{fig:utk_cdf}
\end{figure}

\textbf{How do \FL Solutions Compare to \ERM?}
\Cref{fig:utk_cdf} shows the CDFs and CVaRs for the UTKFace classification task with $\epsilon=0.02$. 
In training, both \FL and \RFL outperform \ERM across the entire loss spectrum, with higher CDFs and lower CVaRs. In validation, the methods demonstrate similar performance, with overlapping curves.

\begin{figure}[t!]
    \centering
    \includegraphics{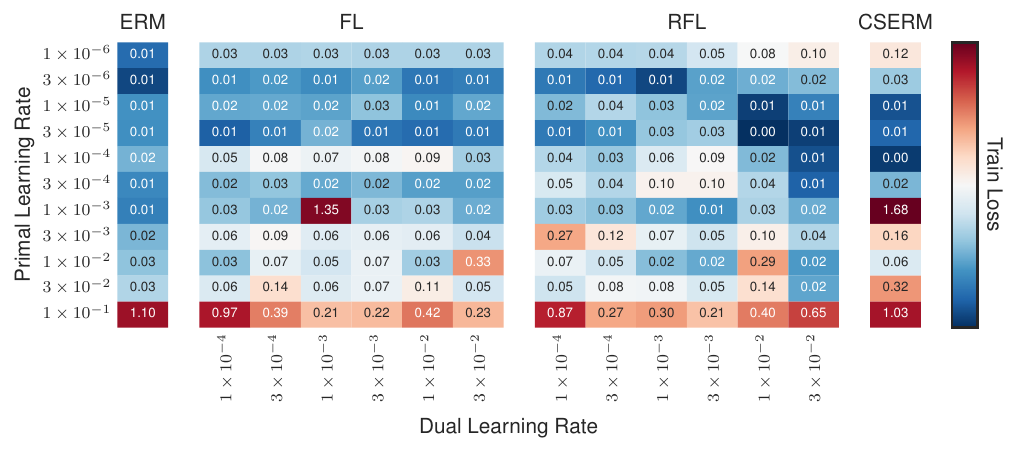}
    \caption{Average training losses at the end of training for UTKFace age regression. \textbf{\FL and \RFL demonstrate good performance across various combinations of primal and dual learning rates.} \captioncomment{$\epsilon=0.02$.}}
    \label{fig:utk_robustness}
\end{figure}

\textbf{Hyper-parameter Robustness.}
\Cref{fig:utk_robustness} displays the training losses at the end of training for the UTKFace age regression task (with $\epsilon=0.02$). Both \FL and \RFL exhibit strong performance across a wide range of primal and dual learning rate combinations, covering multiple orders of magnitude. This suggests that \FL and \RFL are relatively robust to these hyperparameters.

\newpage
\subsection{Direct Preference Optimization}
\label{app:llms}

\textbf{Can we Learn with Feasible Learning?}
\Cref{table:LLMs} shows the performance at the end of training for the direct preference optimization task for Llama and Zephyr models. We report the DPO loss of \ERM and \RFL for both models across training and test splits.

For both models, \RFL results in a slight increase in the average train and test losses compared to \ERM. However, \RFL also significantly reduces the maximum losses in both the training and testing phases.

\begin{table}[t!]
    \centering
    \caption{DPO losses at the end of training for the Llama and Zephyr models on the direct preference optimization task. 
    \textbf{\RFL generally leads to reduced maximum losses, suggesting a more compact loss distribution, while \ERM achieves better average performance.}}
    \label{table:LLMs}
    \begin{tabular}{lccccc}
        \toprule
        \multirow{2}{*}[-0.5\dimexpr \aboverulesep + \belowrulesep + \cmidrulewidth]{\textbf{Model}} & \multirow{2}{*}[-0.5\dimexpr \aboverulesep + \belowrulesep + \cmidrulewidth]{\textbf{Method}} & \multicolumn{2}{c}{\textbf{Train}} & \multicolumn{2}{c}{\textbf{Test}} \\
        \cmidrule(lr){3-4} \cmidrule(lr){5-6}
        & & \multicolumn{1}{c}{Avg. Loss} & Max Loss & \multicolumn{1}{c}{Avg. Loss} & Max Loss \\ 
        \midrule
        LLaMA      & \ERM      & 0.387 & 3.769 & 0.395 & 3.395 \\
        LLaMA      & \RFL       & 0.401 & 2.843 & 0.424 & 2.693 \\
        \midrule
        Zephyr     & \ERM      & 0.269 & 3.354 & 0.402 & 7.423 \\
        Zephyr     & \RFL       & 0.321 & 1.368 & 0.457  & 5.350 \\
        \bottomrule
    \end{tabular}
\end{table}

\textbf{How does Resilience Help?}
Due to the high computational cost of fine-tuning these billion-parameter models, we did not perform a resilience impact study in the DPO task. Additionally, we did not assess the hyper-parameter robustness of \FL and \RFL.

Our main objective is to demonstrate the effectiveness of \RFL in direct preference optimization, focusing on its ability to shape the loss CDFs. Our results prove that \RFL can effectively mitigate high losses in preference optimization, even with minimal hyper-parameter tuning.

\begin{figure}[t!]
\centering
\begin{subfigure}{.5\textwidth}
  \centering
  \includegraphics{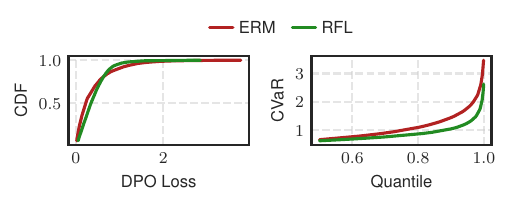}
  \caption{Train CDF and CVaR.}
  \label{fig:llama_cdf_train}
\end{subfigure}%
\begin{subfigure}{.5\textwidth}
  \centering
  \includegraphics{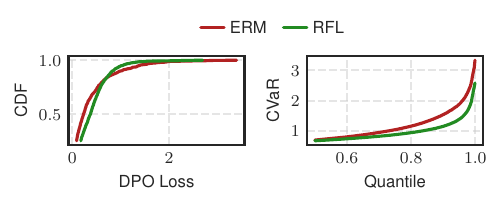}
  \caption{Test CDF and CVaR.}
  \label{fig:llama_cdf_test}
\end{subfigure}
\caption{Llama 3.1-8B train and test empirical CDF and CVaR. \captioncomment{The test plots in (b) correspond to \cref{fig:cdf_shaping} in \S\ref{exps:robustness}.}}
\label{fig:llama_cdf}
\end{figure}

\begin{figure}[t!]
\centering
\begin{subfigure}{.5\textwidth}
  \centering
  \includegraphics{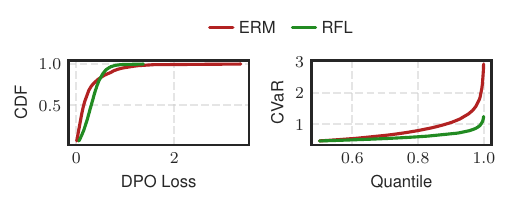}
  \caption{(a) Train CDF and CVaR.}
  \label{fig:zephyr_cdf_train}
\end{subfigure}%
\begin{subfigure}{.5\textwidth}
  \centering
  \includegraphics{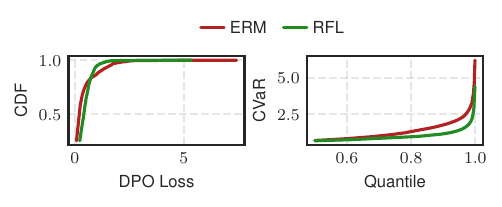}
  \caption{(b) Test CDF and CVaR. }
  \label{fig:zephyr_cdf_test}
\end{subfigure}
\caption{Zephyr 3B train and test empirical CDF and CVaR. The plots illustrate the differences in loss distributions between \ERM and \FL, with \FL providing a more compact and stable loss distribution, especially in the tails.}
\label{fig:zephyr_cdf}
\end{figure}

\textbf{How do \FL Solutions Compare to \ERM?}
\Cref{table:LLMs} also shows that \FL reduces the maximum losses both in training and test splits, highlighting its ability to reduce worst-case losses. \Cref{fig:llama_cdf,fig:zephyr_cdf} display the CDFs and CVaRs for the train and test sets for the Llama and Zephyr models, respectively. For both models, we observe that \FL yields a more compact distribution of losses compared to \ERM, with higher CDF values and lower CVaRs in the tail regions, indicating a reduced risk of large losses.

\subsubsection*{Supplementary References}

\bibliographystylesup{bib_style}
\bibliographysup{sup}

\newpage

\end{document}